\documentclass{article}

\usepackage[preprint]{neurips_2025}

\setcitestyle{authoryear, open={(},close={)}}

\usepackage[utf8]{inputenc} 
\usepackage[T1]{fontenc}    
\usepackage{hyperref}       
\usepackage{url}            
\usepackage{booktabs}       
\usepackage{amsfonts}       
\usepackage{nicefrac}       
\usepackage{microtype}      
\usepackage{xcolor}         

\usepackage{threeparttable}
\usepackage{amsmath}
\usepackage{amssymb}
\usepackage{multirow}
\usepackage{makecell}
\usepackage{graphicx}
\usepackage{enumitem}
\usepackage{wrapfig}

\title{Availability-aware Sensor Fusion \\via Unified Canonical Space}

\author{%
  Dong-Hee Paek \quad\quad Seung-Hyun Kong\thanks{corresponding author}\\
  CCS Graduate School of Mobility\\
  KAIST\\
  \texttt{\{donghee.paek, skong\}@kaist.ac.kr} \\
}

\begin{document}

\maketitle

\begin{abstract}
Sensor fusion of camera, LiDAR, and 4-dimensional (4D) Radar has brought a significant performance improvement in autonomous driving. However, there still exist fundamental challenges: deeply coupled fusion methods assume continuous sensor availability, making them vulnerable to sensor degradation and failure, whereas sensor-wise cross-attention fusion methods struggle with computational cost and unified feature representation. This paper presents availability-aware sensor fusion (ASF), a novel method that employs unified canonical projection (UCP) to enable consistency in all sensor features for fusion and cross-attention across sensors along patches (CASAP) to enhance robustness of sensor fusion against sensor degradation and failure. As a result, the proposed ASF shows a superior object detection performance to the existing state-of-the-art fusion methods under various weather and sensor degradation (or failure)  conditions. Extensive experiments on the K-Radar dataset demonstrate that ASF achieves improvements of 9.7\% in $AP_{BEV}$ (87.2\%) and 20.1\% in $AP_{3D}$ (73.6\%) in object detection at IoU=0.5, while requiring a low computational cost. All codes are available at \url{https://github.com/kaist-avelab/k-radar}.
\end{abstract}

\section{Introduction}
\label{sec:intro}

Autonomous driving technology has advanced rapidly, with multiple companies adopting multi-sensor fusion approaches that combine two or more sensors, such as cameras, LiDAR, and 4-dimensional (4D) Radar, to achieve more robust and reliable perception \citep{self_driving_survey,sensor_fusion_survey}. Cameras uniquely provide color information but struggle with depth estimation; LiDAR delivers high-resolution 3-dimensional (3D) point cloud data but is less reliable in adverse weather conditions \citep{adverse_weahter_survey}; and 4D Radar, despite its relatively low angular resolution, offers robustness in adverse weather and directly measures relative velocity \citep{kradar,vod,rtnhp,survey_4d_radar}. This complementary nature initiated sensor fusion between camera, LiDAR, and 4D Radar to improve perception performance and reliability compared to a single-sensor \citep{cmt,bevfusion_nips,bevfusion_icra,rcfusion}.

\begin{figure*}[t]
\centering
\includegraphics[width=1.0\textwidth]{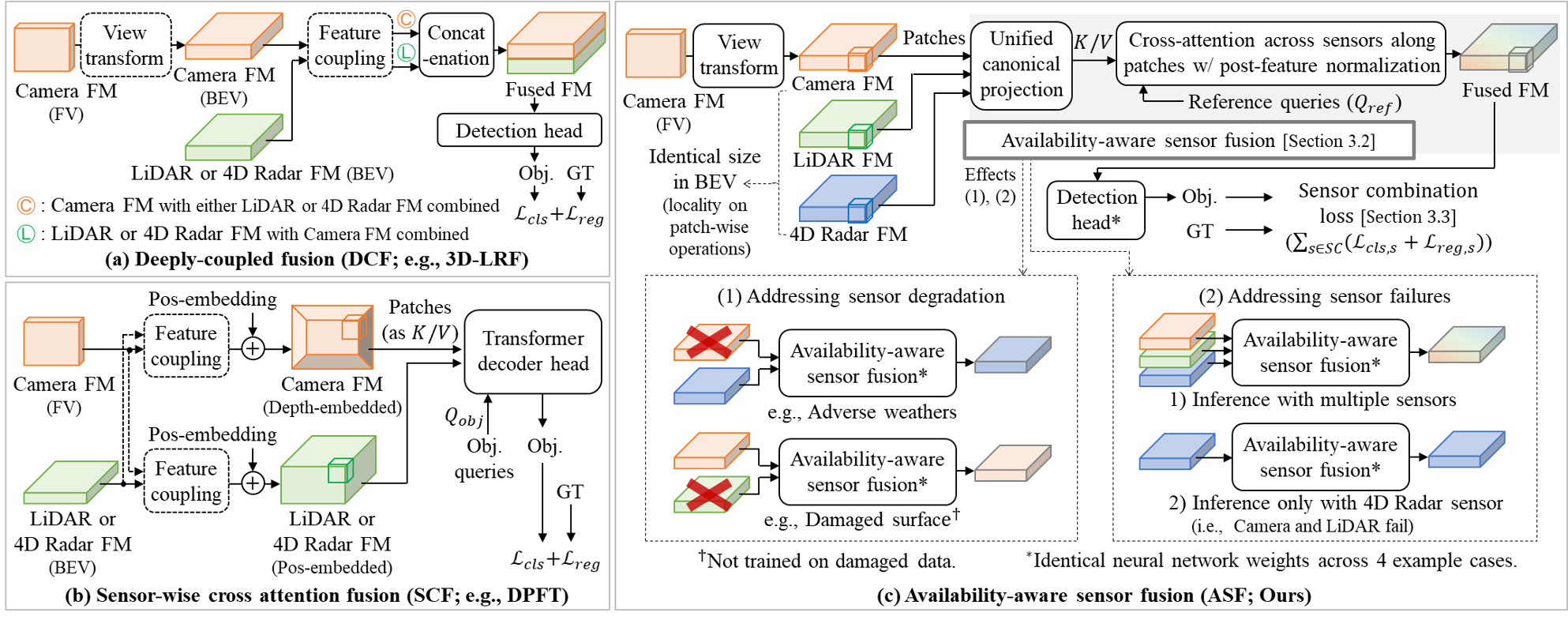}
\caption{{Comparison of sensor fusion methods: (a) DCF (e.g., 3D-LRF \citep{lrf}), (b) SCF (e.g., CMT \citep{cmt}), and (c) ASF.} FV, BEV, Obj., GT, $\mathcal{L}_{cls}$ and $\mathcal{L}_{reg}$ stand for `front-view', `bird's eye-view', `objects', `ground truths', `classification loss', and `regression loss', respectively. `Feature coupling' refers to methods that combine features from multiple sensors to create new features. Optional components are in dashed lines; for example, \citep{pointpainting} combines camera and LiDAR features without transforming the camera viewpoint, while \citep{cmt} fuses features through a transformer decoder head \citep{detr} without explicit feature coupling. ASF does not apply feature coupling to ensure independence between sensors.}
\label{fig:structure_comparison}
\end{figure*}

Most multi-modal sensor fusion methods can be categorized into two methods. The first is deeply coupled fusion (DCF), which directly combines feature maps (FMs) extracted by sensor-specific encoder, as illustrated in Fig.~\ref{fig:structure_comparison}-(a). While it is simple to implement and computationally efficient with excellent performance in various benchmarks \citep{lrf,bevfusion_icra,bevfusion_nips,simplebev,nuscenes,kitti}, it assumes all sensors are functioning properly and consistently. This makes DCF vulnerable to sensor degradation due to adverse weathers, surface-damages, and sensor failure. Moreover, DCF requires retraining the entire neural network when the number of sensors changes, as the size of the fused FM (i.e., the input to the detection head) changes. The second method is sensor-wise cross-attention fusion (SCF), which divides features extracted from each sensor into patches with positional-embedding (e.g., depth information for camera \citep{petr,detr3d}) and selectively combines available patches using cross-attention (Fig.~\ref{fig:structure_comparison}-(b)), allowing it to handle cases where some sensors are degraded. However, SCF does not have sensor scalability, since the method does not project sensor-specific features into a standardized representation \citep{unified_segmentation,unified_vision_language,unified_image_text}. In addition, SCF incurs computational complexity that scales with the number of patches, resulting in substantial computational overhead when processing multiple sensors with numerous patches \citep{dpft,cmt,transfusion}.

\begin{figure}[t]
\centering
\includegraphics[width=0.89\textwidth]{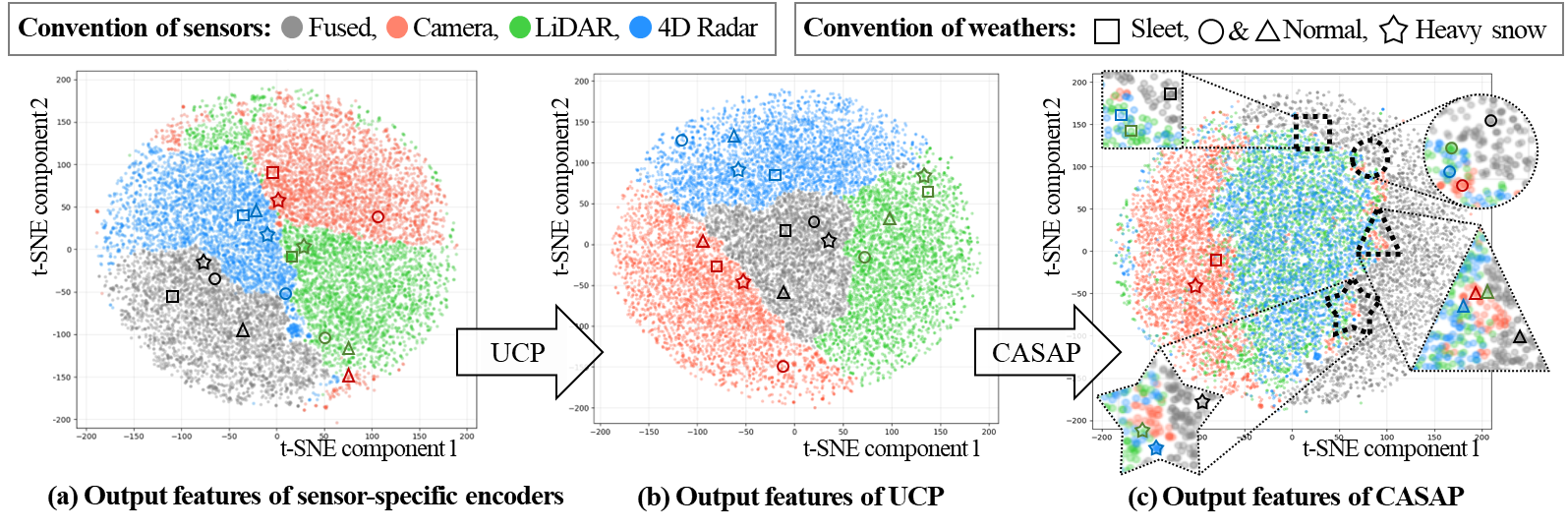}
\caption{{Visualization of feature representation with t-SNE \citep{tsne} at different stages of ASF for `Sedan' class.} Red, green, blue, and gray dots represent features from camera, LiDAR, 4D Radar, and fused features, respectively. Symbols in solid lines such as circle and triangle, square, and star indicate normal, sleet, and heavy snow conditions, respectively. (a) Initial output features from sensor-specific encoders show inconsistent distribution across sensors. (b) After unified canonical projection (UCP), features become better aligned to the fused feature. (c) After cross-attention across sensors along patches (CASAP), features from available sensors form cohesive clusters (in dashed symbols) based on weather conditions. Note that in adverse weather, camera features show larger deviation due to the degradation. Additional visualizations are in Appendix B.}
\label{fig:tsne_visualization}
\end{figure}

One of the fundamental limitations of existing fusion methods stems from inconsistencies in feature representation across different sensors \citep{mobileye,sensor_fusion_survey_inconsistence}. Cameras produce 2D RGB images, whereas LiDAR generates 3D point clouds, and 4D Radar produces low-resolution tensors with power values. Therefore, each sensor extracts features with different representations for the same object, making consistent fusion challenging (as shown in Fig.~\ref{fig:tsne_visualization}-(a)). To address this inconsistency, an ideal strategy could project features from different sensors into a unified canonical space for fusion. The concept of `True Redundancy \citep{mobileye}', that ensures sensor independence while maintaining canonical feature representation for any sensor combination, suggests a promising direction for highly reliable and robust sensor fusion. 

Therefore, we propose availability-aware sensor fusion (ASF) method (Fig.~\ref{fig:structure_comparison}-(c)), in which each sensor performs independently while being complementarily fused through a projection to a unified canonical space. As a result, the proposed method addresses the limitations of both DCF and SCF simultaneously. The key innovation of ASF is in two sub-modules; First, unified canonical projection (UCP) projects features from each sensor into a unified space based on common criteria (i.e., canonical). Since UCP is optimized using the same reference query for all sensors, inconsistencies in sensor features are eliminated. While Fig.~\ref{fig:tsne_visualization}-(a) shows sensor features represented without clear patterns, Fig.~\ref{fig:tsne_visualization}-(b) demonstrates how UCP aligns the features from each sensor to the fused feature. Second, cross-attention across sensors along patches (CASAP) estimates the availability of sensors through patch-wise cross-attention on features projected into the unified canonical space, assigning higher weights to features from available sensors and lower weights to features from missing or degraded sensors. Unlike SCF that applies the cross-attention across all sensors ($N_s$) and patches ($N_p$) (i.e., $O(N_qN_sN_p)$ for $N_q$ queries), ASF only applies the cross-attention across sensors along patches (i.e., $O(N_qN_s)$). Because of this, ASF eliminates complex positional-embedding and improves computationally efficiency. Additionally, it applies normalization to ensure that (camera, LiDAR, and 4D Radar) features can be processed consistently by the detection head regardless of sensor combination. In contrast to Fig.~\ref{fig:tsne_visualization}-(b), where each sensor's features remain separate, Fig.~\ref{fig:tsne_visualization}-(c) shows how sensor features cluster together after CASAP (except for camera become useless in adverse weathers, as shown in Fig.~\ref{fig:qualitative_results}). This enables ASF to flexibly handle sensor degradation or failure and to achieve the reliability embodied in the `True Redundancy' concept \citep{mobileye} for autonomous driving implementation. 

To integrate the availability awareness into the detection head, we propose a sensor combination loss (SCL) that optimizes learning outcomes across all sensor combinations. SCL considers individual sensor unavailability during the training, enabling the system to maintain high performance in the presence of unexpected sensor failure or adverse weather conditions. The effectiveness of our proposed ASF method has been validated on the K-Radar dataset \citep{kradar}, demonstrating improvements of 9.7\% in $AP_{BEV}$ (87.2\%) and 20.1\% in $AP_{3D}$ (73.6\%) for detection performance at IoU=0.5 compared to state-of-the-art (SOTA) methods \citep{lrf,l4dr}, which includes the performance in extreme situations such as sensor degradation or failure (i.e., unavailable).

The main contributions of this paper are as follows:
(1) We propose ASF based on UCP and CASAP that achieves superior performance to SOTA methods and robust performance against sensor degradation and failure. 
(2) We propose SCL for the loss function to optimize the detection performance for all possible sensor combinations.
(3) Through extensive experiments on the K-Radar dataset, we demonstrate that ASF achieves the high performance with low computational load.

The remainder of this paper is organized as follows: Section \ref{sec:related} introduces existing methods, and Section \ref{sec:proposed} describes the proposed ASF in detail. Section \ref{sec:experiments} presents experimental settings and results on the K-Radar dataset, analyzing performance in various scenarios including sensor degradation and failure. Section \ref{sec:conclusion} concludes the paper with a summary and discusses the limitations. All codes and logs for ASF are available at \url{https://github.com/kaist-avelab/k-radar}.

\section{Related Works}
\label{sec:related}

\subsection{Deeply Coupled Fusion (DCF)}
DCF constructs a fused feature map (FM) by concatenating FMs from each sensor. Most studies focus on fusing camera with LiDAR or 4D Radar \citep{bevfusion_icra,bevfusion_nips,rcfusion,lxl}, or combining LiDAR and 4D Radar \citep{lrf,l4dr}. Implementations range from directly fusing the front-view (FV) camera image with LiDAR or 4D Radar without view transformation \citep{pointpainting} to applying learnable BEV transforms \citep{lss} and concatenating at the BEV stage \citep{bevfusion_icra,bevfusion_nips,rcfusion}. DCF method is straightforward to implement and computationally efficient compared to SCF, demonstrating strong performance across multiple benchmarks.

Recent DCF studies have improved performance by applying feature coupling, where the FMs of each sensor are enhanced with FMs of other sensors using multi-layer perception (MLP) or attention mechanisms \citep{attention}. 3D-LRF \citep{lrf} demonstrated superior performance to the conventional DCFs \citep{bevfusion_nips,bevfusion_icra} by applying attention between LiDAR and 4D Radar voxel features before concatenation. L4DR \citep{l4dr} achieved SOTA performance on the K-Radar dataset by weighting each sensor's FM using coupled BEV FMs with LiDAR and 4D Radar. However, DCF assumes all sensors are functioning normally, as they rely on the fused FM constructed by concatenating FMs from all sensors, making DCF vulnerable to sensor degradation and failure. This limitation arises because the training process does not expose the model to potential sensor degradation or failure scenarios that commonly occur during deployment.

\subsection{Sensor-wise Cross-attention Fusion (SCF)}
SCF divides sensor-specific FMs into patches and dynamically combines them through cross-attention in a transformer decoder head \citep{cmt}, inherently accommodating sensor availability. TransFusion \citep{transfusion} is the first SCF that addresses sensor availability, but its sequential fusion method (e.g., performing LiDAR detection first and then fusing with camera data) makes inference impossible when LiDAR is unavailable. CMT \citep{cmt} presents an availability-aware sensor fusion of camera and LiDAR data using a transformer decoder head without applying feature coupling to individual sensors. However, without feature coupling for camera, CMT relies on positional embedding to incorporate depth information into camera FMs \citep{petr}, which results in 3D patches ($H\! \times\! W \!\times\! D$). This leads to computational complexity of $O(N_qN_sN_p)$ (where $N_q$, $N_s$, and $N_p$ are the number of queries, sensors, and patches, respectively), causing explosive growth in computational cost and memory usage. For instance, CMT requires 8 A100 GPUs with 80GB VRAM to train with a batch size of 16.

Recently, DPFT \citep{dpft} creates independent FMs and projects sensor-agnostic query points onto different FMs to verify sensor availability, achieving 56.1\% $AP_{3D}$ at IoU=0.3 (using only 4D Radar and camera). Unlike methods that utilize entire FMs, DPFT achieves reasonable computational efficiency by employing deformable attention \citep{deformable_attention} that considers varying receptive fields using only a small number of key points. However, similar to CMT, DPFT performs object detection using variable object queries, which does not establish a common representation across different sensors (as illustrated in Fig.~\ref{fig:tsne_visualization}).

\section{Proposed Methods}
\label{sec:proposed}
\subsection{Sensor Fusion Framework}

The overall sensor fusion framework consists of three stages: (1) sensor-specific encoders (i.e., backbones) that extract same-sized bird's eye-view (BEV) feature maps (FMs) from each sensor data (i.e., RGB image, LiDAR point cloud, and 4D Radar tensor), (2) the proposed availability-aware sensor fusion (ASF) network that is described in following subsection \ref{sec:asf}, and (3) a detection head that detects objects from the fused FM.

Focusing on our ASF contribution, we utilize established methods for the sensor-specific encoders and detection head. Specifically, we adopt BEVDepth \citep{bevdepth}, SECOND \citep{second}, and RTNH \citep{kradar} backbones for camera, LiDAR, and 4D Radar, respectively, along with a SSD detection head \citep{ssd}. Further specifications regarding the overall sensor fusion framework, such as sensor-specific encoders and detection head, are provided in Appendix A.

\subsection{Availability-aware Sensor Fusion (ASF)}
\label{sec:asf}

As illustrated in Fig.~\ref{fig:structure_comparison}-(c), the proposed ASF consists of two key components: unified canonical projection (UCP) and cross-attention across sensors along patches (CASAP).

\noindent\textbf{Unified Canonical Projection (UCP).} One of the key challenge in multi-modal sensor fusion is the inherent inconsistency of features from different sensors \citep{mobileye,sensor_fusion_survey_inconsistence}, as visualized in Fig.~\ref{fig:tsne_visualization}-(a). To tackle this, we divide BEV FMs into an equal number of patches for all sensors and train projection functions to transform features from each sensor into a unified space based on the same criteria (i.e., reference query in CASAP). To formally define our methods, we first represent the same-sized FMs of each sensor as:
\begin{equation}
\mathrm{\mathbf{FM}}^s\in\mathbb{R}^{C_s{\times}H{\times}W}, s \in \{ S_C, S_L, S_R \},
\end{equation}
\noindent where $C_s$ denotes the channel dimension for sensor $s$, $H$ and $W$ represent the identical height and width of BEV FMs for all sensors, respectively, and $S_C$, $S_L$, and $S_R$ denote camera, LiDAR, and 4D Radar sensors, respectively. Each FM is then divided into patches $\mathbf{F}_{p}^s$ with height $P_H$ and width $P_W$:
\begin{equation}
\mathcal{T}_p(\mathbf{FM}^s)\!=\!\{\mathbf{F}_{p,i}^s|\mathbf{F}_{p,i}^s \!\in\! \mathbb{R}^{C_s\times P_H \times P_W}, i\!=\!1\!:\!N_p\},
\end{equation}
\noindent where $\mathcal{T}_p(\cdot)$ is the operation that divides each FM into patches, $N_p\!=\!(H/P_H)\!\times\! (W/P_W)$ is the number of patches, which is identical across all sensors since each FM has the same spatial size, and `$1\!:\!N_p$' denotes `$1,2,...,N_p$'. Note that since the patches are already spatially aligned (i.e., $\mathbf{F}_{p,i}^{S_C}$, $\mathbf{F}_{p,i}^{S_L}$, and $\mathbf{F}_{p,i}^{S_R}$ correspond to the same position), our method eliminates the use of computationally expensive positional-embedding \citep{petr} required for SCF \citep{cmt,dpft}. Then, we apply a parallel operation along patches that projects each patch to have the same channel dimension $C_u$. This is the UCP operation $\mathcal{U}^s(\cdot)$ that transforms sensor-specific patches into patches in a unified canonical space. The UCP-processed patch $\mathcal{P}_{u}^s$ for each sensor is expressed as:
\begin{align}
&\mathcal{P}^s_u \!=\! \{\mathbf{F}_{u,i}^s|\mathbf{F}_{u,i}^s \!=\!\mathcal{U}^s(\mathrm{LN}(\mathbf{F}_{p,i}^s)) \!\in\! \mathbb{R}^{C_{u}}, i\!=\!1\!:\!N_p\}\\
&\mathcal{U}^s(\cdot)\!=\!\mathrm{LN}({\mathrm{Proj}^{(n_{u})}(\cdot)}), \mathrm{Proj(\cdot)\!=\!\mathrm{GeLU}(\mathrm{MLP}(\cdot)}),
\end{align}
\noindent where LN and $n_{u}$ denote layer normalization \citep{layernorm} for training stability and the number of sequential projection functions incorporating MLP for transformation and GeLU \citep{gelu} for non-linearity, respectively. While our framework allows for repetition of the projection function to increase non-linearity, with 1 or 2 repetitions being sufficient (we use $n_u=2$, which aligns features as demonstrated in Fig.~\ref{fig:tsne_visualization}-(b)). Note that $\mathcal{U}^s$ is trained separately for each sensor based on reference query, which results in alignment of features from all sensors with respect to the fused feature as shown in Fig.~\ref{fig:tsne_visualization}-(b).

\noindent\textbf{Cross-attention Across Sensors Along Patches (CASAP).} The patches $\mathbf{F}_{u}^s$ projected into the unified canonical space by UCP serve as keys ($K$) and values ($V$) for a trainable reference query $\mathbf{Q}_{ref}\!\!\in\! \mathbb{R}^{N_q \times C_u}$ (where $N_q$ is the number of queries $Q$), and we perform cross-attention across sensors along patches as:
\begin{equation}
\label{eq:asf}
\mathbf{Q}_{ref,i}'\!=\!\mathrm{CrossAttn}(Q\!=\!\mathbf{Q}_{ref},K\&V\!\in\!\{\mathbf{F}_{u,i}^{S_C},\mathbf{F}_{u,i}^{S_L},\mathbf{F}_{u,i}^{S_R}\}),i\!=\!1:N_p, \end{equation}
\noindent where $\mathbf{Q}_{ref,i}'$ is the output of the cross-attention applied across sensors for the $i$-th patch. Since $\mathbf{Q}_{ref}$ is trained primarily on features that are mostly available in the training data, it naturally develops high correlation (i.e., high attention scores) with patches from available sensors after the training. Consequently, during inference, $\mathbf{Q}_{ref,i}'$ is predominantly composed of available $\mathbf{F}_{u}^{s}$. The number of heads in cross-attention is a hyper-parameter whose impact is analyzed in subsection \ref{sec:ablations}.

\begin{figure*}[t]
\centering
\includegraphics[width=1.0\textwidth]{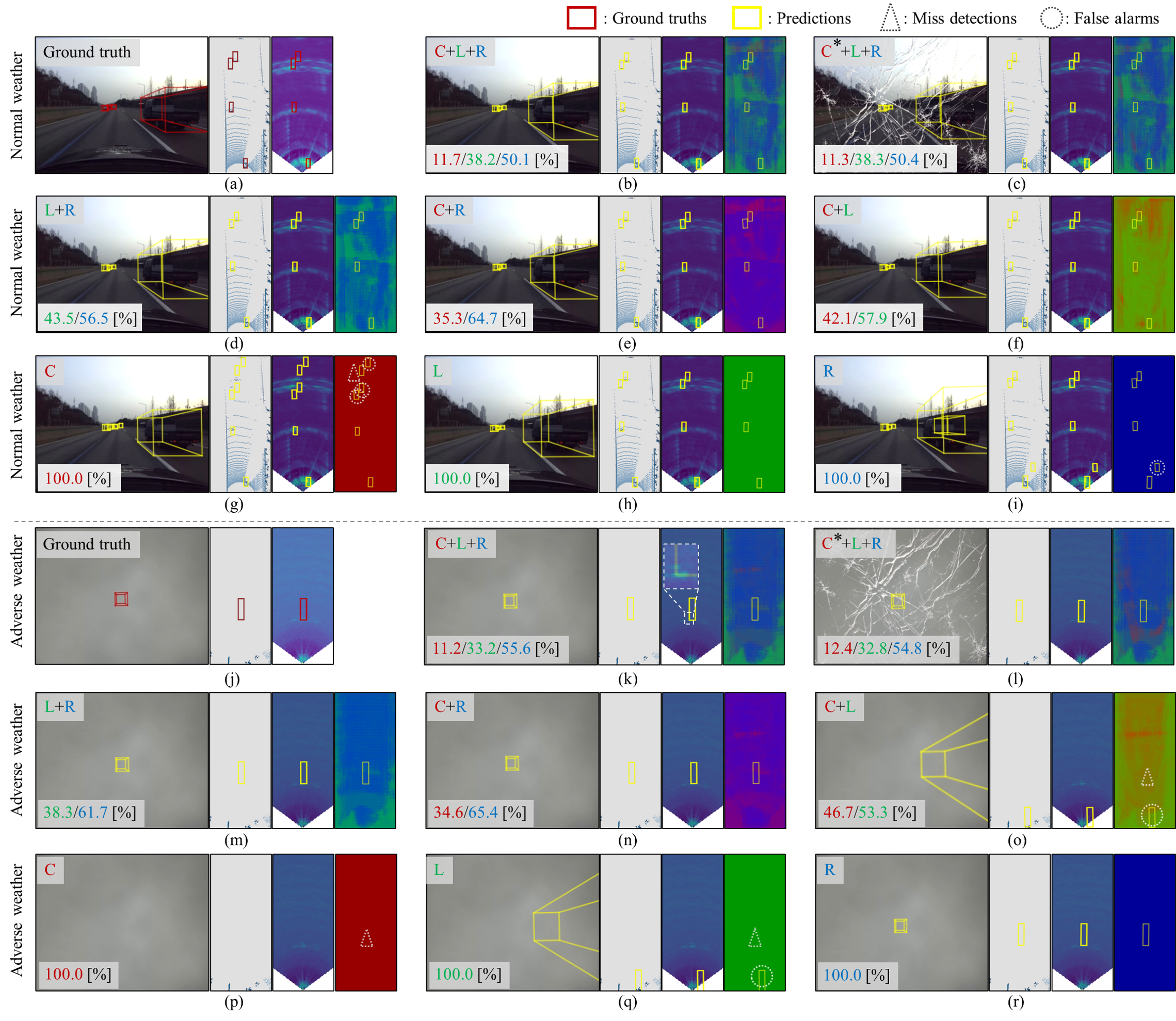}
\caption{{Qualitative results of ASF for various sensor combinations.} We show results for normal and adverse weather conditions in (a-i) and (j-r), respectively, where employed sensors are noted in the top-left corner (C: Camera, L: LiDAR, R: 4D Radar, C*: damaged camera). Each subplot visualizes front-view camera image, LiDAR point cloud, 4D Radar tensor, and a sensor attention map (SAM) showing attention score distribution from cross-attention in CASAP. In the SAMs, red, green, and blue represent attention scores for Camera, LiDAR, and 4D Radar, respectively. For example, a predominantly blue SAM indicates that 4D Radar receives the highest attention scores, meaning that 4D Radar is used primarily for detection in the scene. The bottom-left corner of each subplot shows the proportion of attention scores in colored percentages (\textcolor{red}{C}/\textcolor{green}{L}/\textcolor{blue}{R}[\%]). Note that predictions are visualized on all sensor data, even when a sensor is not employed for detection (e.g., predictions from L+R are also visualized on the camera image).}
\label{fig:qualitative_results}
\end{figure*}

Compared to ASF, cross-attention in SCF is performed across all patches with respect to object queries $\mathbf{Q}_{obj}$ in the transformer decoder head. This can be mathematically expressed as:
\begin{equation}
\label{eq:scf}
\mathbf{Q}_{obj}'\!\!=\!\!\mathrm{CrossAttn}^{(n_{td})}\!(Q\!=\!\mathbf{Q}_{obj}, K\&V\!\!\in\!\!\{\mathbf{F}_{pe,1}^{S_C}\!,\!...,\!\mathbf{F}_{pe,N_p}^{S_C}\!,\!\mathbf{F}_{pe,1}^{S_L}\!,\!...,\!\mathbf{F}_{pe,N_p}^{S_L}\!,\!\mathbf{F}_{pe,1}^{S_R}\!,\!...,\!\mathbf{F}_{pe,N_p}^{S_R}\!\}\!),
\end{equation}
\noindent where $\mathbf{F}_{pe}^{s}$ represents patches with positional-embedding, and $n_{td}$ (usually larger than 6 \citep{petr,cmt}) is the number of stacked transformer decoders. Eq.~\ref{eq:scf} shows the cross-attention across all sensors and all patches (i.e., the $K\&V$ set contains $N_sN_p$ patches, resulting in $O(N_qN_sN_p)$ computational complexity for $N_q$ queries). In contrast, in Eq.~\ref{eq:asf}, the cross-attention is applied across sensors and along patches, which requires only $O(N_qN_s)$ computational operations. This is a significant computational costs reduction as $N_s \ll N_p$. Moreover, as demonstrated in Fig.~\ref{fig:qualitative_results} and Tab.~\ref{tab:main_comparison}, ASF achieves better performance with only a single cross-attention layer than SCF utilizing stacked cross-attention layers (i.e., $n_{td}\!\geq\!6$).

Sequentially, ASF applies post-feature normalization (PN) $\mathcal{N}$ that has a similar structure to $\mathcal{U}^s(\cdot)$ with LN, to ensure that features can be processed consistently by the detection head regardless of the sensor combination. Therefore, PN enables camera, LiDAR, and 4D Radar features for the same object to be consistent. The set of patches $\mathcal{P}_n$ with PN is formulated as:
\begin{align}
&\mathcal{P}_n \!=\!\{\mathbf{F}_{n,i}|\mathbf{F}_{n,i}\!=\!\mathcal{N}(\mathrm{LN}(\mathbf{Q}_{ref,i}'))\!\in\!\mathbb{R}^{C_u}\!,i\!=\!1\!:\!N_p\}\\
&\mathcal{N}(\cdot)\!=\!\mathrm{LN}({\mathrm{Proj}^{(n_{n})}(\cdot)}), \mathrm{Proj(\cdot)\!=\!\mathrm{GeLU}(\mathrm{MLP}(\cdot)}),
\end{align}
\noindent where $n_{n}$ denotes the number of sequential projections, with 1 or 2 being sufficient to increase non-linearity. Unlike Fig.~\ref{fig:tsne_visualization}-(b) where features from different sensors occupy distinct regions, Fig.~\ref{fig:tsne_visualization}-(c) illustrates how PN causes sensor features to converge into unified clusters.

\begin{table*}[t]
\small
\caption{{Performance comparison of 3D object detection on K-Radar \citep{kradar} benchmark v1.0.} C, L, and R represent Camera, LiDAR, and 4D Radar, respectively. The \textbf{bold} and \underline{underlined} values indicate the best and the second-best, respectively. Note that the two ASF models for sensors L+R and C+L+R share the same neural network weights trained with C+L+R, which means that ASF for L+R represents a scenario where camera becomes unavailable. Nor., Ove., Sle., L.s., and H.s. denote `Normal', `Overcast', `Sleet', `Light snow', and `Heavy snow', respectively.}
\label{tab:main_comparison}
\centering
\begin{tabular}{c|c|c|c|c|ccccccc}
\hline\hline
Methods & Sensors & IoU & \multicolumn{1}{c|}{Metric} & Total & Nor. & Ove. & Fog & Rain & Sle. & L.s. & H.s. \\
\hline
\multirow{4}{*}{\makecell{RTNH \\(\citeauthor{kradar})}} & \multirow{4}{*}{R} & \multirow{2}{*}{0.3} & BEV & 41.1 & 41.0 & 44.6 & 45.4 & 32.9 & 50.6 & 81.5 & 56.3 \\
 &  &  & 3D & 37.4 & 37.6 & 42.0 & 41.2 & 29.2 & 49.1 & 63.9 & 43.1 \\
\cline{3-12}
 &  & \multirow{2}{*}{0.5} & BEV & 36.0 & 35.8 & 41.9 & 44.8 & 30.2 & 34.5 & 63.9 & 55.1 \\
 &  &  & 3D & 14.1 & 19.7 & 20.5 & 15.9 & 13.0 & 13.5 & 21.0 & 6.36 \\
\hline
\multirow{4}{*}{\makecell{RTNH \\(\citeauthor{kradar})}} & \multirow{4}{*}{L} & \multirow{2}{*}{0.3} & BEV & 76.5 & 76.5 & 88.2 & 86.3 & 77.3 & 55.3 & 81.1 & 59.5 \\
 &  &  & 3D & 72.7 & 73.1 & 76.5 & 84.8 & 64.5 & 53.4 & 80.3 & 52.9 \\
\cline{3-12}
 &  & \multirow{2}{*}{0.5} & BEV & 66.3 & 65.4 & 87.4 & 83.8 & 73.7 & 48.8 & 78.5 & 48.1 \\
 &  &  & 3D & 37.8 & 39.8 & 46.3 & 59.8 & 28.2 & 31.4 & 50.7 & 24.6 \\
\hline
\multirow{4}{*}{\makecell{3D-LRF \\(\citeauthor{lrf})}} & \multirow{4}{*}{L+R} & \multirow{2}{*}{0.3} & BEV & {84.0} & {83.7} & {89.2} & {95.4} & {78.3} & {60.7} & {88.9} & {74.9} \\
 &  &  & 3D & {74.8} & {81.2} & {87.2} & {86.1} & {73.8} & {49.5} & {87.9} & {67.2} \\
\cline{3-12}
 &  & \multirow{2}{*}{0.5} & BEV & {73.6} & {72.3} & {88.4} & {86.6} & {76.6} & {47.5} & {79.6} & {64.1} \\
 &  &  & 3D & {45.2} & {45.3} & {55.8} & {51.8} & {38.3} & {23.4} & {60.2} & {36.9} \\
 \hline
 \multirow{4}{*}{\makecell{L4DR \\(\citeauthor{l4dr})}} & \multirow{4}{*}{L+R} & \multirow{2}{*}{0.3} & BEV & {79.5} & {86.0} & {89.6} & {89.9} & {81.1} & {62.3} & {89.1} & {61.3} \\
 &  &  & 3D & {78.0} & {77.7} & {80.0} & {88.6} & {79.2} & {60.1} & {78.9} & {51.9} \\
\cline{3-12}
 &  & \multirow{2}{*}{0.5} & BEV & {77.5} & {76.8} & {88.6} & {89.7} & {78.2} & {59.3} & {80.9} & {53.8} \\
 &  &  & 3D & {53.5} & {53.0} & {64.1} & {73.2} & {53.8} & {46.2} & {52.4} & {37.0} \\
  \hline
 \multirow{8}{*}{\makecell{ASF \\(Proposed)}} & \multirow{4}{*}{L+R} & \multirow{2}{*}{0.3} & BEV
 & {\underline{88.6}} & {\underline{88.1}} & {\textbf{90.3}} & {\textbf{99.0}} & {\textbf{89.1}} & {\underline{80.4}} & {\textbf{89.4}} & {\textbf{78.7}} \\
 &  &  & 3D
 & {\underline{87.3}} & {\underline{86.6}} & {\underline{89.8}} & {\underline{90.7}} & {\textbf{88.6}} & {\underline{80.0}} & {\textbf{88.8}} & {\textbf{77.5}} \\
\cline{3-12}
 &  & \multirow{2}{*}{0.5} & BEV
 & {\underline{87.0}} & {\underline{86.2}} & {\textbf{90.2}} & {\underline{90.8}} & {\textbf{88.8}} & {\underline{78.2}} & {\textbf{88.6}} & {\textbf{71.0}} \\
 &  &  & 3D
 & {\underline{72.9}} & {\underline{64.6}} & {\underline{86.6}} & {\textbf{79.6}} & {\textbf{73.4}} & {\textbf{70.0}} & {\underline{77.6}} & {\textbf{66.7}} \\
  \cline{2-12}
 & \multirow{4}{*}{C+L+R} & \multirow{2}{*}{0.3} & BEV
 & {\textbf{88.6}} & {\textbf{88.2}} & {\underline{90.2}} & {\underline{98.9}} & {\underline{89.0}} & {\textbf{80.4}} & {\underline{89.2}} & {\underline{78.4}} \\
 &  &  & 3D
 & {\textbf{87.4}} & {\textbf{87.0}} & {\textbf{90.1}} & {\textbf{90.7}} & {\underline{88.2}} & {\textbf{80.0}} & {\underline{88.6}} & {\underline{77.4}} \\
\cline{3-12}
 &  & \multirow{2}{*}{0.5} & BEV
 & {\textbf{87.2}} & {\textbf{86.7}} & {\underline{90.1}} & {\textbf{90.8}} & {\underline{88.7}} & {\textbf{78.3}} & {\underline{88.3}} & {\underline{70.9}} \\
 &  &  & 3D
 & {\textbf{73.6}} & {\textbf{71.8}} & {\textbf{87.0}} & {\underline{79.4}} & {\underline{73.0}} & {\underline{67.5}} & {\textbf{78.0}} & {\underline{66.4}} \\
\hline\hline
\end{tabular}
\end{table*}

\begin{wraptable}{R}{0.5\columnwidth}
\small
\caption{{Comparison of VRAM and FPS evaluated on the K-Radar benchmark v1.0.} The \textbf{bold} and \underline{underlined} values indicate the best and the second-best, respectively. The unit of VRAM and FPS are GB and Hz, respectively.}
\label{tab:computational_efficiency}
\centering
\begin{tabular}{c|c|c|c}
\hline\hline
Methods & Sensors & VRAM & FPS \\
\hline
3D-LRF (\citeauthor{lrf}) & L+R & \textbf{1.2} & 5.04 \\
DPFT (\citeauthor{dpft}) & C+R & 4.0 & 11.5 \\
\hline
ASF & L+R & \underline{1.5} & \textbf{20.5} \\
(Proposed) & C+L+R & 1.6 & \underline{13.5} \\
\hline\hline
\end{tabular}
\end{wraptable}

Finally, to transform $\mathcal{P}_n$ back to the original BEV size ($H\!\times\!W$), we apply a reshape operation $\mathcal{T}_n(\cdot)$. Since the size of $\mathcal{P}_n$ is $C_u\!\times\! N_p \!=\! C_u\!\times\! N_H\!\times\! N_W \!=\! C_u\!\times\! (H/P_H)\!\times\!(W/P_W)$, the fused FM $\mathbf{FM}_{\mathrm{fused}}$ can be obtained as:
\begin{align}
\mathbf{FM}_{\mathrm{fused}} \!=\! \mathcal{T}_n(\mathcal{P}_n \!\in\! \mathbb{R}^{C_u\times N_p})\!\in\! \mathbb{R}^{C_{q}\times H\times W},
\end{align}
\noindent where $C_{q}$ is the quotient of $C_u/(P_H \!\times\! P_W)$ as we design $C_u \!=\! P_H\!\times\! P_W \!\times\! C_{q}$. Since the resulting channel dimension $C_{q}$ may be insufficient for containing feature representation due to reduced channel dimensions after reshaping, in the implementation, we increase the number of patches by a factor of $n_p$ (i.e., $N_p \!\!=\!\! N_H\!\times\! N_W \!\!\rightarrow\!\! N_p \!=\! n_p\!\!\times\!\! N_H\!\!\times\!\! N_W$). Consequently, the channel dimension of $\mathbf{FM}_{\mathrm{fused}}$ increases from $C_{q}$ to $n_p\!\times\! C_{q}$, and impact of this modification is evaluated in subsection \ref{sec:ablations}.

\subsection{Sensor Combination Loss (SCL)}
Leveraging the consistent size of $\mathbf{FM}_{\mathrm{fused}}$ (which serves as the input to the detection head) regardless of sensor combinations, we propose an SCL that enables simultaneous optimization across multiple sensor configurations. The proposed SCL is formalized as:
\begin{equation}
\mathcal{L}_{SCL} \!=\! \sum_{s \in \mathcal{SC}} (\mathcal{L}_{cls,s} \!+\! \mathcal{L}_{reg,s}),
\end{equation}
\noindent where $\mathcal{SC}$ represents the set of $7$ possible sensor combinations ($S_C$-only, $S_L$-only, $S_R$-only, $S_L\!+\!S_R$, $S_C\!+\!S_R$, $S_C\!+\!S_L$, and $S_C\!+\!S_L\!+\!S_R$), where $S_C$, $S_L$, and $S_R$ denote camera, LiDAR, and 4D Radar, respectively. $\mathcal{L}_{cls,s}$ and $\mathcal{L}_{reg,s}$ are the classification and regression losses for each sensor combination. SCL explicitly prepares for the potential sensor unavailability by optimizing across all sensor combinations in the training, enabling the model to recognize that available sensors perform better than others (e.g., 4D Radar outperform camera in adverse weather). As demonstrated in Tab.~\ref{tab:ablations}, SCL enhances the performance of the proposed ASF method when compared to ASF without SCL.

\section{Experiments}
\label{sec:experiments}

\subsection{Experimental Setup}
\label{sec:exp_setup}

\noindent\textbf{Dataset and Metrics.}
K-Radar \citep{kradar} is a large-scale autonomous driving dataset with a broad range of conditions including time (day, night), weather (normal, rain, fog, snow, sleet), road types (urban, highway, mountain), and sensors (4D Radar, LiDAR, camera, GPS). Notably, K-Radar is the only dataset with data captured in adverse weather conditions. 

For comparison with SOTA methods, we utilize two K-Radar benchmark variants. Benchmark v1.0 \citep{kradar,lrf,l4dr} focuses on the `Sedan' class within a driving corridor region of [0m, 72m] × [-6.4m, 6.4m] × [-2m, 6m] (X×Y×Z). For ablation studies and qualitative analysis, we use benchmark v2.0, which covers a wider area [0m, 72m] × [-16m, 16m] × [-2m, 7.6m] and includes both `Sedan' and `Bus or Truck' classes. We evaluate using $AP_{3D}$ and $AP_{BEV}$ at IoU thresholds of 0.3 and 0.5, while also reporting VRAM usage and FPS based on the same hardware setup.

\begin{wraptable}{R}{0.56\columnwidth}
    \caption{{Performance comparison of ASF under various sensor combinations on K-Radar \citep{kradar} benchmark v2.0.} We indicate the employed sensors (C: Camera, L: LiDAR, R: 4D Radar) and report $AP_{3D}$ at IoU = 0.3 for `Sedan' and `Bus or Truck' classes. `Nor.', `Ove.', 'Sle.` and `H.s.' refer to `Normal', `Overcast', `Sleet', and `Heavy snow', respectively. $*$ denotes the sensor unavailability, as shown in Fig.~\ref{fig:qualitative_results}. The \textbf{bold} and \underline{underlined} values indicate the best and the second-best, respectively. All ten ASF models share the same neural network weights trained for R+L+C. Performance under additional weather conditions (Fog, Rain, and Light Snow) and with other evaluation metric $AP_{BEV}$ is provided in Appendix D.}
    \label{tab:sensor_failure}
    \centering
    \small
    \begin{tabular}{c|c|c|ccccc}
        \hline\hline
        Class & Sensors & Total & Nor. & Ove. & Sle. & H.s. \\
        \hline
        \multirow{10}{*}{Sedan} & R     & 47.3 & 40.7 & 58.8 & 45.9 & 56.5 \\
                                & L     & 73.0 & 73.0 & 86.1 & 64.9 & 54.5 \\
                                & C     & 14.8 & 14.9 & 7.71 & - & - \\                                                           
                                & C$^{*}$ & 3.71 & 3.72 & 3.17 & - & - \\
                                & L+R   & 77.3 & 77.7 & 87.3 & 74.4 & 65.4 \\
                                & C+R   & 52.7 & 49.1 & 62.4 & 46.0 & 57.2 \\
                                & C+L   & 76.4 & \underline{78.3} & 86.5 & 64.2 & 57.1 \\
                                & C+L+R & \textbf{79.3} & \textbf{78.8} & \underline{87.6} & \underline{74.2} & \textbf{65.8} \\
                                & C$^{*}$+L+R & \underline{77.6} & 78.2 & \textbf{87.7} & \textbf{74.4} & \underline{65.4} \\
                                & C+L$^{*}$+R & 58.9 & 58.8 & 66.6 & 52.3 & 58.2 \\
        \hline
        \multirow{10}{*}{\makecell{Bus\\or\\Truck}} & R     & 34.2 & 22.9 & 40.9 & 21.1 & 51.2 \\
                                       & L     & 54.9 & \textbf{53.7}&74.8&69.1&37.8 \\
                                       & C     & 9.59 & 9.02 & 17.2 & - & - \\
                                       & C$^{*}$ & 3.65 & 3.66 & 0.00 & - & - \\
                                       & L+R   & 59.9 & 52.5 & 71.6 & 68.2 & 68.9 \\
                                       & C+R   & 36.2 & 24.4 & 41.4 & 23.4 & 56.5 \\
                                       & C+L   & 53.0 & 49.1 & 60.1 & \textbf{72.1} & 39.6 \\
                                       & C+L+R & \textbf{60.4} & \underline{52.7} & \textbf{77.4} & 69.2 & \underline{68.9} \\
                                       & C$^{*}$+L+R & \underline{60.1} & 52.1 & \underline{72.0} & \underline{70.9} & \textbf{69.1} \\
                                       & C+L$^{*}$+R & 40.0 & 31.5 & 38.2 & 28.9 & 54.8 \\
        \hline\hline
    \end{tabular}
\end{wraptable}

\noindent\textbf{Implementation Details.}
We implement the ASF on a single RTX3090 GPU with 24GB VRAM. ASF is trained for 11 epochs using AdamW \citep{adamw} optimizer with a learning rate $0.001$ and a batch size $2$. The voxel size for the fused FM is set to $0.4$m, consistent with \citep{kradar}.

\subsection{Comparison of ASF to SOTA Methods}
\label{sec:comparison}

Following the benchmark v1.0 of the K-Radar \citep{kradar}, we compare the proposed ASF with SOTA methods including 3D-LRF \citep{lrf} and L4DR \citep{l4dr}, and we use RTNH \citep{kradar} for single-sensor performance. In addition to the detection performance comparison with DCF methods \citep{lrf,l4dr}, we evaluate computational efficiency against DPFT \citep{dpft} which is the only open-sourced SCF method available for K-Radar.

\noindent\textbf{Detection Performance.} As shown in Tab.~\ref{tab:main_comparison}, ASF significantly outperforms SOTA methods across various weather conditions. Compared to previous SOTA L4DR \citep{l4dr}, ASF achieves substantial improvements of 9.7\% in $AP_{BEV}$ (87.2\% vs. 77.5\%) and 20.1\% in $AP_{3D}$ (73.6\% vs. 53.5\%) at IoU=0.5. These improvements are particularly remarkable in challenging conditions like sleet (67.5\% vs. 46.2\% $AP_{3D}$) and heavy snow (66.4\% vs. 37.0\% $AP_{3D}$). Notably, both ASF configurations (L+R and C+L+R) use identical neural network weights yet maintain comparable performance, demonstrating the system's ability to gracefully handle sensor degradation. Even with only LiDAR and Radar, ASF achieves 87.0\% $AP_{BEV}$ and 72.9\% $AP_{3D}$ at IoU=0.5, nearly matching the full sensor suite's performance.

\noindent\textbf{Computational Efficiency.} ASF achieves exceptional computational efficiency for real-time autonomous driving applications. As demonstrated in Tab.~\ref{tab:computational_efficiency}, ASF with LiDAR and 4D Radar processes at 20.5 Hz, approximately 4× faster than 3D-LRF (5.04 Hz) using identical sensors. Even with all three sensors, ASF maintains 13.5 Hz, exceeding the 10 Hz threshold typically required for autonomous driving systems \citep{adsystem_survey}. This efficiency results from our CASAP, which applies cross-attention across sensors along patches rather than across all sensors and patches. Furthermore, ASF maintains a compact memory footprint (1.5-1.6 GB), comparable to 3D-LRF (1.2 GB) and substantially lower than DPFT (4.0 GB).

\subsection{Addressing Sensor Degradation and Failure}
\label{sec:qualitative_results}

A key advantage of ASF is the robust performance under sensor degradation or failure. As shown in Tab.~\ref{tab:sensor_failure} and Fig.~\ref{fig:qualitative_results}, ASF dynamically adapts to different sensor combinations without retraining. Under normal conditions (Fig.~\ref{fig:qualitative_results}-(a-i)), ASF effectively utilizes all available sensors with attention weights distributed according to each sensor's reliability. However, ASF's true value emerges in challenging scenarios. In adverse weather (Fig.~\ref{fig:qualitative_results}-(j-r)), camera and LiDAR measurements are significantly degraded or disappear completely. In these critical situations, ASF automatically redistributes attention toward the more reliable 4D Radar, as evidenced by the predominant blue coloration in the sensor attention maps (SAMs) and corresponding attention percentages. Even with damaged sensors (denoted by * in Tab.~\ref{fig:qualitative_results} and Fig.~\ref{fig:qualitative_results}), ASF maintains near-optimal performance. For example, with a damaged camera (C*), C*+L+R shows 77.6\% AP$_{3D}$, which is only 1.7\% lower than with fully sensors (79.3\%). This robustness stems from the unified canonical projection (which creates a common feature space) and the cross-attention mechanism (which estimates sensor reliability).

The qualitative results in Fig.~\ref{fig:qualitative_results} demonstrate that in adverse weather, when LiDAR measurements disappear and camera visibility severely degrades, reliable object detection is only possible with active 4D Radar and ASF is fully using 4D Radar. Note that all results shown are from the same ASF model with identical weights, illustrating how ASF dynamically adjusts attention to maintain detection performance across varying sensor availabilities.

\subsection{Ablation Studies}
\label{sec:ablations}

\begin{wraptable}{O}{0.53\columnwidth}
\caption{{Ablation study of ASF.} ASF performance for different components and parameters: $P$ (patch size), $C_u$ (channel dimension in unified canonical space), $n_p$ (number of patches multiplier), $n_h$ (number of heads in CASAP) and SCL, using $AP_{3D}$ at IoU=0.3 for both `Sedan' and `Bus or Truck' on the K-Radar benchmark v2.0.}
\label{tab:ablations}
\small
\centering
\begin{tabular}{c|c|c|c|c|c|cc}
\hline\hline
Exp. & $P$ & $C_u$ & $n_p$ & $n_h$ & SCL & Sedan & Bus \\
\hline
(a) & 5 & 512 & 1 & 8 & & 76.1 & 45.7 \\
(b) & 5 & 512 & 4 & 8 & & 76.4 & 47.4 \\
(c) & 2 & 512 & 4 & 8 & & 77.2 & 49.7 \\
(d) & 2 & 256 & 8 & 8 & & 77.6 & 57.9 \\
(e) & 2 & 256 & 8 & 8 & $\checkmark$ & 79.3 & 58.2 \\
(f) & 2 & 256 & 8 & 16 & & 77.5 & 60.2 \\
(g) & 2 & 256 & 8 & 16 & $\checkmark$ & 79.3 & 60.4 \\
\hline\hline
\end{tabular}
\end{wraptable}

Tab.~\ref{tab:ablations} presents ablation studies of key ASF components, analyzing five factors: patch size ($P$), channel dimension ($C_u$), patches multiplier ($n_p$), number of attention heads ($n_h$), and sensor combination loss (SCL). Our findings reveal that smaller patch sizes ($P\!\!=\!\!2$) improve performance through finer feature extraction, while balancing reduced channel dimension ($C_u\!\!=\!\!256$) with increased patches multiplier ($n_p\!\!=\!\!8$) maintains or enhances results; furthermore, increasing attention heads ($n_h\!=\!16$) benefits `Bus or Truck' detection, and incorporating SCL consistently improves performance across configurations by enhancing robustness to varying sensor availability. The optimal configuration combines $P\!\!=\!\!2$, $C_u\!\!=\!\!256$, $n_p\!\!=\!\!8$, $n_h\!\!=\!\!16$ with SCL.

\section{Conclusion}
\label{sec:conclusion}
This paper introduces availability-aware sensor fusion (ASF), which addresses sensor availability challenges in autonomous driving by transforming features into a unified canonical space through UCP and CASAP. Our approach maintains computational efficiency ($O(N_qN_s)$) while providing robust fusion for sensor degradation or failure. The proposed sensor combination loss further enhances robustness by optimizing across all possible sensor combinations. Experiments on the K-Radar dataset demonstrate significant improvements over SOTA methods (9.7\% in $AP_{BEV}$ and 20.1\% in $AP_{3D}$ at IoU=0.5), with consistent performance across various weather conditions and sensor combinations.

\noindent\textbf{Limitations.} Despite ASF's strong performance, the camera network's capabilities remain a limitation. As shown in Fig.~\ref{fig:qualitative_results}-(g) and (p), camera-based object detection is less precise, particularly in adverse weather. In Fig.~\ref{fig:tsne_visualization}-(c), while LiDAR and 4D Radar features are well integrated, camera features remain more separated in feature space. Enhancing the camera backbone could further boost system performance, especially in favorable weather conditions, where visual information is valuable.

\setcounter{figure}{3}
\setcounter{table}{4}

\appendix

\section*{Appendix}

The appendix is organized as follows. Section~\ref{sec:app_framework} provides detailed explanations of the sensor fusion framework, including specific implementations of sensor-specific encoders and the detection head. Section~\ref{sec:app_tsne} presents additional t-SNE visualizations that demonstrate the effectiveness of the unified canonical space. Section~\ref{sec:app_attention} analyzes attention score distributions across different weather conditions and distance ranges. Section~\ref{sec:app_performance} includes comprehensive performance results across various sensor combinations and alternative stereo camera configurations.

\section{Details of Sensor Fusion Framework}
\label{sec:app_framework}

\begin{wrapfigure}{R}{0.75\columnwidth}
\centering
\includegraphics[width=0.75\textwidth]{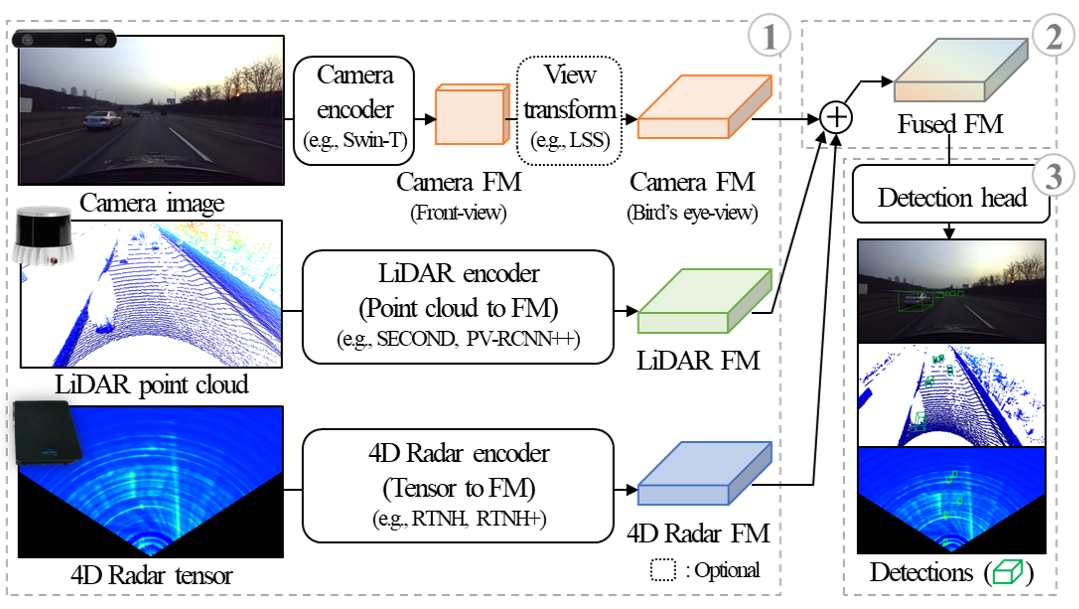}
\caption{{Overall sensor fusion framework for camera, LiDAR, and 4D Radar.} `FM' denotes `feature map'.}
\label{fig:fusion_framework}
\end{wrapfigure}

As shown in Fig.~\ref{fig:fusion_framework}, the sensor fusion framework is divided into three stages: (1) sensor-specific encoders \citep{swin,lss,second,pvrcnn,kradar,rtnhp} that extract feature maps (FMs) from different types of multi-modal sensors (e.g., camera, LiDAR, and 4D Radar), (2) a fusion method that integrates these FMs, and (3) a task-dependent detection head. Stage (1) has evolved independently for each sensor \citep{vit,pointpillars,kradar}, while stage (3) employs a task-dependent method (e.g., YOLO head \citep{yolo} for object detection) regardless of sensor type. Therefore, in this work, we propose a fusion method focusing on stage (2) rather than stages (1) and (3). Details of both existing and proposed fusion methods are illustrated in Fig.~1. In this section, we describe both the sensor-specific encoders and the detection head of the proposed availability-aware sensor fusion (ASF).

\subsection{Sensor-specific Encoders}

Sensor-specific encoders extract FMs from data of each sensor. Because different sensors provide data in different modalities—cameras produce 2D RGB images, LiDAR generates 3D point clouds, and 4D Radar outputs low-resolution tensors with power values—research on these encoders has evolved independently for each sensor type.

For ASF, we adopt established backbone architectures for each sensor and pre-train these backbones before ASF training. Each encoder is trained for 11 epochs following the K-Radar \citep{kradar} training protocol. During ASF training, we freeze all sensor-specific encoders to ensure that only the ASF components and detection head are learned. Specifically, `UCP' and `CASAP' are the ASF components updated during ASF training, which denote \emph{unified canonical projection} and \emph{cross-attention across sensors along patches}, respectively. Details for each encoder are as follows:

\noindent\textbf{Camera.}  
Camera-based sensor encoders for autonomous driving typically require transforming front-view (FV) images into a bird’s-eye-view (BEV) format for 3D object detection. Current approaches can be broadly categorized into two methods: (1) learnable BEV transformations such as Lift-Splat-Shoot (LSS) \citep{lss,bevfusion_icra,bevfusion_nips}, or (2) maintaining FV while using transformer decoders \citep{petr} that incorporate depth information as positional embeddings in BEV space. Although methods like PETR \citep{petr} have improved 3D object detection by injecting depth embeddings, they inevitably increase computational complexity due to the additional depth dimension. These two approaches form the foundation of camera-based fusion \citep{lxl,rcfusion,bevfusion_icra,bevfusion_nips,cmt}, commonly seen in deeply coupled fusion (DCF) and sensor-wise cross-attention fusion (SCF).

To develop a computationally efficient method, we opt for the LSS approach and adopt BEVDepth \citep{bevdepth}, which supervises the LSS depth estimation using LiDAR point clouds. For encoding images, we use the Swin-S model of SwinTransformer as the 2D backbone, achieving 83\% top-1 accuracy on ImageNet \citep{imagenet}. We resize front camera images to $704 \!\times\! 256$ pixels for encoding. The encoded 2D FMs are projected into BEV FMs via LSS with a grid size of 0.2\,m. To match the 0.4\,m grid size used in K-Radar, we add a convolutional layer with stride 2. During training, we employ the depth information projected from LiDAR point clouds onto the image and, inspired by \citep{mono_distil}, we use the LiDAR backbone’s (SECOND) BEV FMs for distillation. As shown in Fig.~3, camera images have limited utility in adverse weather, so we train the camera encoder only on normal or overcast scenes (sequences 1–20) from K-Radar.

A limitation of this approach is that promising stereo or multi-view detection models cannot be fully leveraged, because the K-Radar dataset \citep{kradar}—the only one offering camera, LiDAR, and 4D Radar data in adverse weather—provides object annotations only for the front view. Moreover, unlike the KITTI dataset with its longer stereo baseline of 0.54\,m, K-Radar uses waterproof ZED cameras with a baseline of just 0.12\,m, limiting stereo-based detection for distant objects. Consequently, we employ a monocular camera method despite inherent performance constraints.

\noindent\textbf{LiDAR.}  
LiDAR provides precise depth information in the form of dense point clouds, often used as a reference for labeling other sensors \citep{kradar,auto_label}. We adopt the SECOND \citep{second} model as the LiDAR encoder, which is open-source in K-Radar and widely used in various benchmarks. Leveraging LiDAR’s high spatial resolution, we set the voxel size to 0.05\,m and apply multiple layers of sparse convolution to generate BEV FMs with a 0.4\,m grid size.

\noindent\textbf{4D Radar.}  
Unlike LiDAR, which uses Time-of-Flight (ToF) measurement, Radar detects spatial information through signal processing, which can introduce sidelobe noise. Traditional 3D Radar typically filters out noisy measurements to produce a sparse point cloud focusing on object presence \citep{radar_tutorial}. In contrast, 4D Radar captures additional elevation information, yielding higher-density point clouds \citep{dual}. This property is exemplified by K-Radar \citep{kradar,enhanced_kradar}. We adopt RTNH \citep{kradar}, the baseline provided in K-Radar, as our 4D Radar encoder, following the original network settings and design.

\subsection{Detection Head}

Following the K-Radar baseline, we use an SSD detection head \citep{ssd,complexyolo} for the 3D object detection task. Our implementation uses two anchor boxes per object class, oriented at $0^\circ$ and $90^\circ$. Each anchor box is parametrized by eight values: the 3D center coordinates ($x_c$, $y_c$, $z_c$), 3D size ($x_l$, $y_l$, $z_l$), and orientation ($\cos(\text{yaw})$, $\sin(\text{yaw})$), as in \citep{complexyolo}.

To address the severe class imbalance in object detection (negative samples far outnumber positives), we use the focal loss \citep{focal} for classification. For bounding box regression, we use smooth L1 loss to measure differences between predictions and ground truth, improving training stability. During inference, we select the class with the highest logit score for each proposal, then apply non-maximum suppression to remove redundant overlapping detections, producing the final predictions.

\section{t-SNE Visualizations}
\label{sec:app_tsne}

\subsection{Additional t-SNE Visualizations}

\begin{figure*}[h]
\centering
\includegraphics[width=1.0\textwidth]{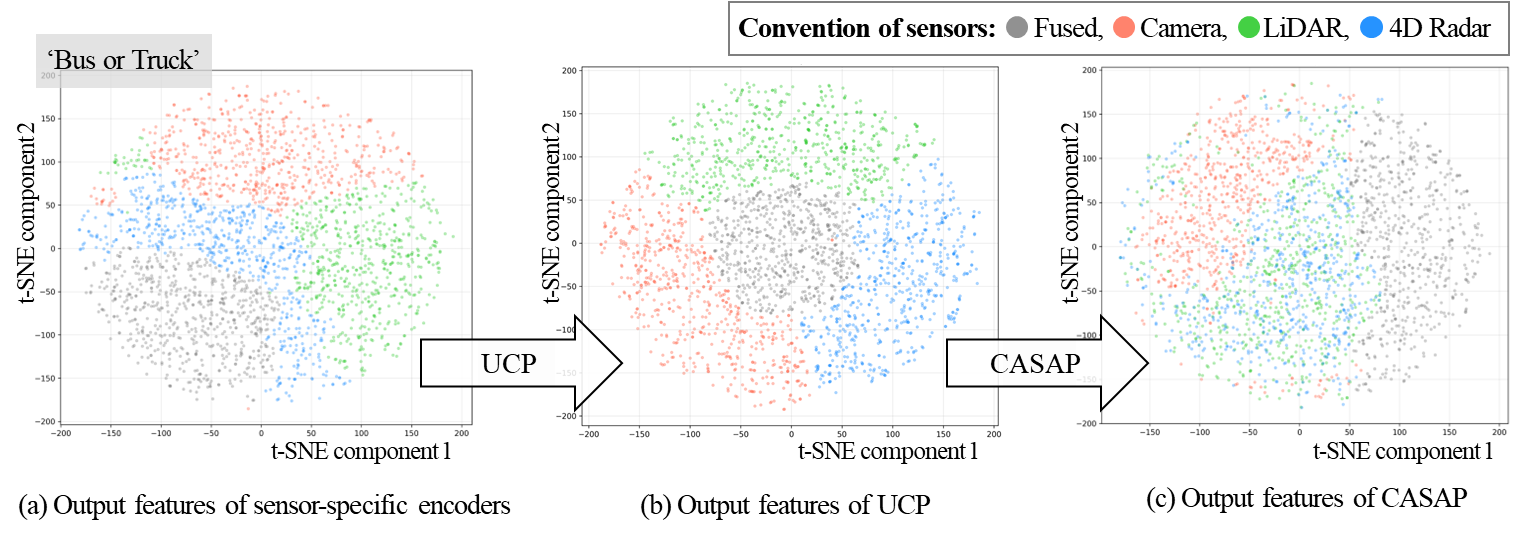}
\caption{{Visualization of feature representations through t-SNE \citep{tsne} at different stages of the proposed ASF for the `Bus or Truck' class.} `UCP' and `CASAP' denote \emph{unified canonical projection} and \emph{cross-attention across sensors along patches}, which are the two main components of the proposed ASF.}
\label{fig:tsne_appendix}
\end{figure*}

This section provides additional feature visualizations to complement those in Fig.~2, which only shows results for the `Sedan' class. Fig.~\ref{fig:tsne_appendix} presents t-SNE \citep{tsne} visualizations of the `Bus or Truck' class at different stages of the ASF. These visualizations further demonstrate how ASF unifies feature representations from different sensors.

Consistent with Fig.~2, we observe three main trends in the t-SNE visualizations. First, the initial features from the sensor-specific encoders are scattered without clear alignment across different sensors. Next, after UCP, sensor-specific features become better aligned with the fused features. Finally, following CASAP, features from available sensors form cohesive clusters (in this case, LiDAR and 4D Radar).

These findings underscore the effectiveness of ASF in creating a unified canonical space for multi-sensor features. Note that in K-Radar, the `Sedan' class has 37,213 instances, whereas the `Bus or Truck' class has 7,449 instances, leading to sparser data in Fig.~\ref{fig:tsne_appendix} compared to Fig.~2.

\subsection{Details of t-SNE Visualizations}

To visualize object-specific features in each sensor’s feature map (FM), we use RoIAlign \citep{roialign} to crop features within bounding boxes. RoIAlign extracts fixed-size FMs, making it suitable for t-SNE (which requires consistent dimensions). We then visualize these cropped features in the unified canonical space. For those extracted before projection (where channel dimensions may differ), we downsample as needed to ensure consistent feature dimensions.

\section{Analysis of Attention Scores across Sensors}
\label{sec:app_attention}

\begin{table}[h]
\centering
\small
\caption{Attention scores ratios [\%] across different weather conditions}
\begin{tabular}{c|ccccccc}
\hline\hline
Sensor & Normal & Overcast & Fog & Rain & Sleet & Light Snow & Heavy Snow \\
\hline
Camera & 11.6 & 11.1 & 11.3 & 11.1 & 11.1 & 10.9 & 11.0 \\
LiDAR & 39.5 & 37.5 & 37.5 & 37.3 & 35.9 & 36.3 & 34.7 \\
4D Radar & 48.9 & 51.4 & 51.2 & 51.6 & 53.0 & 52.8 & 54.3 \\
\hline\hline
\end{tabular}
\label{tab:app_att_weather}
\end{table}

\begin{table}[h]
\centering
\small
\caption{Attention scores ratios [\%] across different distance ranges}
\begin{tabular}{c|ccccccccc}
\hline\hline
Sensor & 0-8m & 8-16m & 16-24m & 24-32m & 32-40m & 40-48m & 48-56m & 56-64m & 64-72m \\
\hline
Camera & 9.8 & 10.1 & 10.4 & 10.9 & 11.1 & 11.3 & 11.9 & 12.7 & 13.4 \\
LiDAR & 48.4 & 40.2 & 37.2 & 36.0 & 34.9 & 34.8 & 35.0 & 35.5 & 36.0 \\
4D Radar & 41.8 & 49.7 & 52.4 & 53.1 & 54.0 & 53.9 & 53.1 & 51.8 & 50.6 \\
\hline\hline
\end{tabular}
\label{tab:app_att_distance}
\end{table}

Tables~\ref{tab:app_att_weather} and \ref{tab:app_att_distance} show the attention score ratios for each sensor according to weather conditions and distance from the ego-vehicle. Intuitively, a higher attention score ratio indicates greater utilization of the corresponding sensor in ASF. Note that the sensor attention maps (SAMs) in Fig.~3 visualize the attention scores for each sensor.

As shown in Tab.~\ref{tab:app_att_weather}, the attention score ratios of camera and LiDAR decrease under severe adverse weather conditions such as sleet and heavy snow (camera: 11.6\% to 11.0\%, LiDAR: 39.5\% to 34.7\%), while 4D Radar shows the opposite trend, increasing from 48.9\% to 54.3\%. This demonstrates that 4D Radar is more robust than camera and LiDAR in adverse weather conditions and provides available measurements when other sensors degrade.

Furthermore, as shown in Tab.~\ref{tab:app_att_distance}, LiDAR's attention score ratio decreases as distance increases (from 48.4\% at 0-8m to 36.0\% at 64-72m), while camera's attention score increases (from 9.8\% to 13.4\%). This indicates that ASF relies more on camera measurements at longer ranges where LiDAR becomes sparse. In contrast, 4D Radar maintains a relatively consistent attention ratio (41.8\% to 50.6\%) regardless of distance. Notably, 4D Radar shows the highest absolute attention scores compared to camera and LiDAR across most conditions. While camera lacks depth information and LiDAR provides very sparse measurements in 3D space, 4D Radar leverages dense tensor data, which enables more reliable feature extraction for sensor fusion.

\section{Additional Detection Results}
\label{sec:app_performance}

\begin{figure*}[h]
\centering
\includegraphics[width=1.0\textwidth]{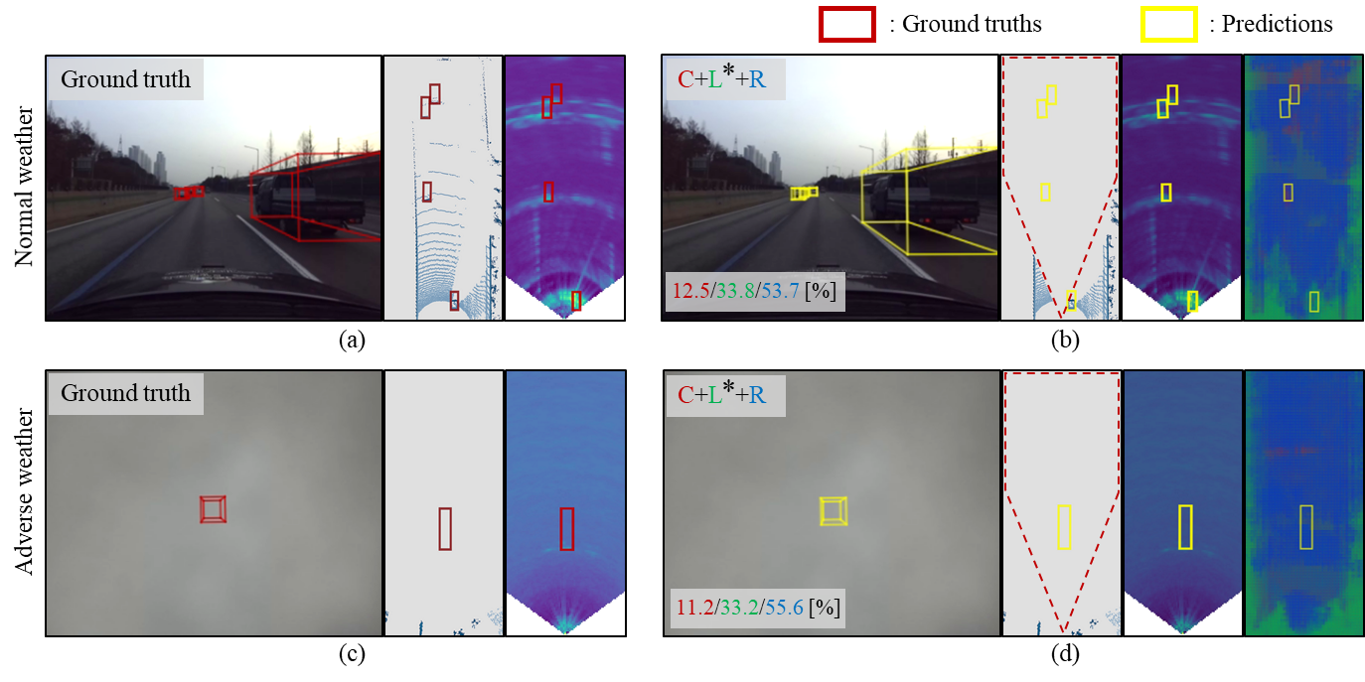}
\caption{{Qualitative results of ASF with C+L$^*$+R (C: Camera, L*: damaged LiDAR, R: 4D Radar).} We show results under (a–b) normal and (c–d) adverse weather conditions. Each grid shows the camera image, LiDAR point cloud, 4D Radar tensor, and a sensor attention map (SAM) illustrating attention scores from CASAP (except for (a) and (c), which only show sensor data for ground truth visualization). In the SAM, red, green, and blue represent attention to camera, LiDAR, and 4D Radar, respectively. For example, a predominantly blue SAM indicates that 4D Radar has the highest attention in that scene. The bottom-left corner of each grid shows attention score proportions (\textcolor{red}{C}/\textcolor{green}{L}/\textcolor{blue}{R}[\%]). Yellow and red boxes denote predictions and ground truth, respectively.}
\label{fig:qualitative_results_appendix}
\end{figure*}

This section presents additional detection results for ASF. First, we provide qualitative results demonstrating ASF's robustness when the LiDAR sensor is damaged, shown in the same road environment as Fig.~3, along with additional results from diverse road scenarios. Second, we demonstrate the robustness of ASF (C+L+R) with respect to random seeds by presenting experimental results on the K-Radar \citep{kradar} benchmark v1.0 using 10 different random seeds (0 to 8 and 2025), complementing Tab.~1 which reports results only for seed 2025. Third, we evaluate alternative camera backbone architectures, including stereo camera-based methods, to assess the impact of improved camera encoders on the overall fusion performance. Finally, we provide comprehensive quantitative performance on K-Radar benchmark v2.0, including $AP_{3D}$ and $AP_{BEV}$ for both `Sedan' and `Bus or Truck' classes at IoU=0.3 and 0.5, complementing Tab.~3 which reports only $AP_{3D}$ at IoU=0.3.

\begin{figure*}[t]
\centering
\includegraphics[width=0.95\textwidth]{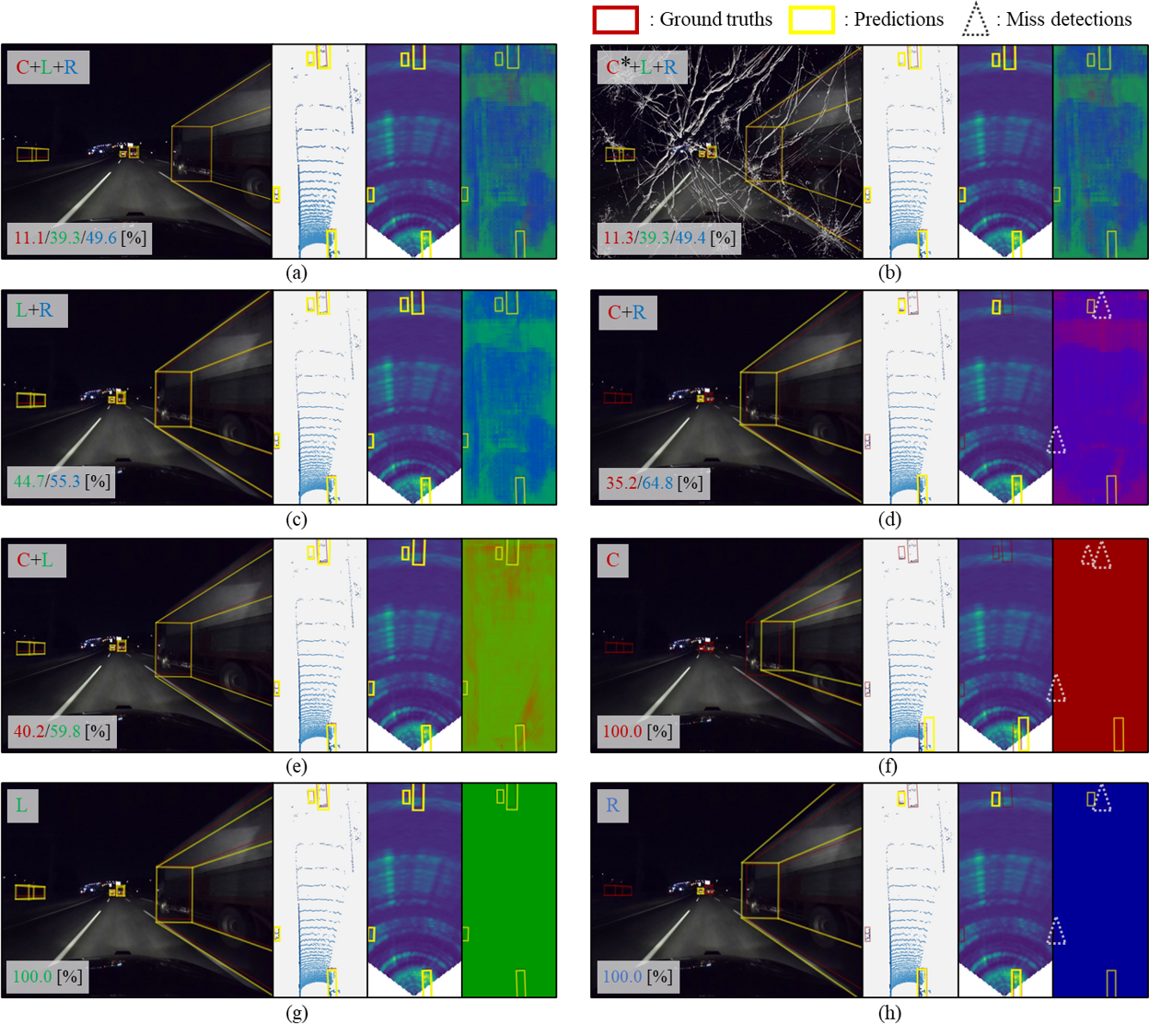}
\caption{{Qualitative results of ASF for various sensor combinations.} We show results low-light nighttime roads. Each grid shows the camera image, LiDAR point cloud, 4D Radar tensor, and a SAM illustrating attention scores from CASAP. In the SAM, red, green, and blue represent attention to camera, LiDAR, and 4D Radar, respectively. The bottom-left corner of each grid shows attention score proportions (\textcolor{red}{C}/\textcolor{green}{L}/\textcolor{blue}{R}[\%]). Yellow and red boxes denote predictions and ground truth, respectively, which are displayed in all panels. Dotted triangles indicate miss detections and are marked in the SAM. Note that predictions are visualized on all sensor data, even when a sensor is not employed for detection (e.g., predictions from L+R are also visualized on the camera image).}
\label{fig:qualitative_results_appendix_1}
\end{figure*}

\begin{figure*}[t]
\centering
\includegraphics[width=0.95\textwidth]{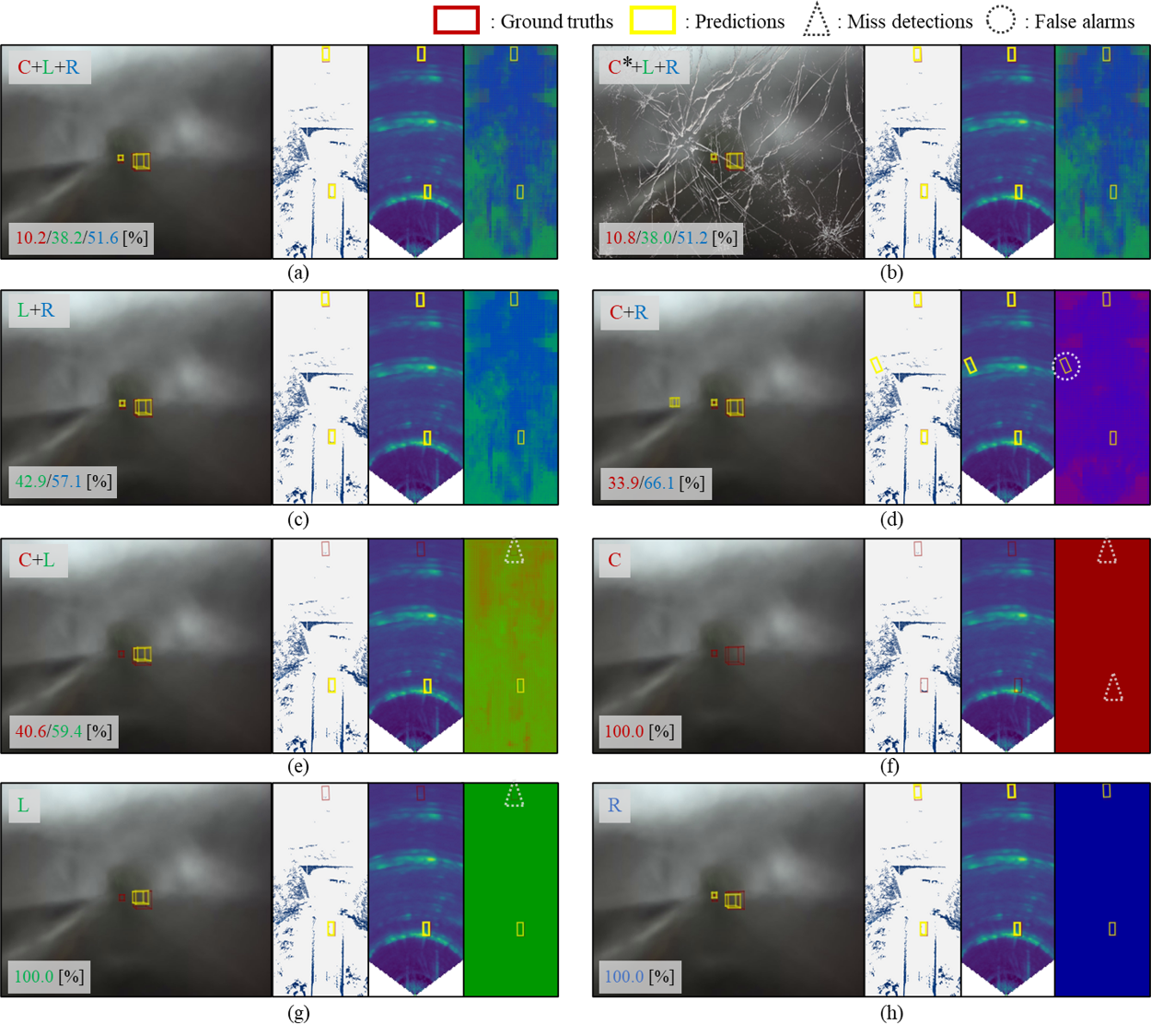}
\caption{{Qualitative results of ASF for various sensor combinations.} We show results under adverse weather conditions. Each grid shows the camera image, LiDAR point cloud, 4D Radar tensor, and a SAM illustrating attention scores from CASAP. In the SAM, red, green, and blue represent attention to camera, LiDAR, and 4D Radar, respectively. The bottom-left corner of each grid shows attention score proportions (\textcolor{red}{C}/\textcolor{green}{L}/\textcolor{blue}{R}[\%]). Yellow and red boxes denote predictions and ground truth, respectively, which are displayed in all panels. Dotted triangles and circles indicate miss detections and false alarms, respectively, and are marked in the SAM. Note that predictions are visualized on all sensor data, even when a sensor is not employed for detection (e.g., predictions from L+R are also visualized on the camera image).}
\label{fig:qualitative_results_appendix_2}
\end{figure*}

\subsection{Additional Qualitative Results}

\noindent\textbf{Damaged LiDAR.} When LiDAR surfaces are damaged or contaminated, LiDAR measurements can be lost \citep{lidar_contamination}. In adverse conditions such as heavy snow, accumulation on the LiDAR sensor can block forward measurements, as illustrated by the red dashed line in Fig.~\ref{fig:qualitative_results_appendix}-(d), even though vehicles are present. We also simulate total loss of forward LiDAR returns from surface damage (e.g., broken sensor), as seen in Fig.~\ref{fig:qualitative_results_appendix}-(b). Despite missing LiDAR data, ASF still detects all objects using only camera and 4D Radar inputs. As reported in Tab.~3, ASF with C+L$^*$+R often performs similarly or even better than C+R alone. Compared to Fig.~3-(b), the sensor attention map (SAM) in Fig.~\ref{fig:qualitative_results_appendix}-(b) confirms the increased reliance on camera or 4D Radar when LiDAR is unavailable.

\noindent\textbf{Additional Results Visualization.} Fig.~\ref{fig:qualitative_results_appendix_1} shows the inference results for multiple vehicles in low-light road environments. When detection is performed using camera-only in low-illumination conditions, numerous miss detections occur, as shown in Fig.~\ref{fig:qualitative_results_appendix_1}-(f). Additionally, as shown in Fig.~\ref{fig:qualitative_results_appendix_1}-(h), in the case of 4D Radar, one of the distant vehicles aligned side by side is covered by the reflection of another vehicle, resulting in miss detection. Furthermore, the leftmost vehicle experiences miss detection as the radio waves are blocked by the road barrier. In the case of Fig.~\ref{fig:qualitative_results_appendix_1}-(a), which utilizes all of 4D Radar, LiDAR, and camera, stable and accurate detection is possible without miss detections.

Fig.~\ref{fig:qualitative_results_appendix_2} shows the inference results for multiple vehicles in adverse weather conditions. As shown in Fig.~\ref{fig:qualitative_results_appendix_2}-(e), when only LiDAR is utilized, nearby vehicles can be detected due to the presence of measurements, but distant vehicles experience miss detection due to the absence of measurements. In contrast, as shown in Fig.~\ref{fig:qualitative_results_appendix_2}-(h), 4D Radar has measurements regardless of distance, enabling stable detection without miss detections. Similar to Fig.~\ref{fig:qualitative_results_appendix_1}-(a), in the case of Fig.~\ref{fig:qualitative_results_appendix_2}-(a), which utilizes all of 4D Radar, LiDAR, and camera, stable and accurate detection is possible without miss detections and false alarms.

\subsection{Additional Quantitative Results}

\begin{table*}[h]
\caption{Additional performance comparison of ASF (C+L+R) under various weather conditions on the K-Radar \citep{kradar} benchmark v1.0. We report both $AP_{BEV}$ and $AP_{3D}$ at IoU = 0.3 for the `Sedan' class, showing mean and standard deviation across 10 random seeds (0 to 8 and 2025).}
\centering
\small
\begin{tabular}{c|c|ccccccc}
\hline\hline
Metric & Total & Normal & Overcast & Fog & Rain & Sleet & Light snow & Heavy snow \\
\hline
\multirow{2}{*}{$AP_{BEV}$} & 88.64 & 88.24 & 90.29 & 98.97 & 88.93 & 86.24 & 89.25 & 78.44 \\
& $\pm$0.12 & $\pm$0.12 & $\pm$0.09 & $\pm$0.27 & $\pm$0.17 & $\pm$2.85 & $\pm$0.11 & $\pm$0.16 \\
\hline
\multirow{2}{*}{$AP_{3D}$} & 87.48 & 86.96 & 89.94 & 90.67 & 88.28 & 80.46 & 88.92 & 76.03 \\
& $\pm$0.21 & $\pm$0.17 & $\pm$0.20 & $\pm$0.08 & $\pm$0.20 & $\pm$0.21 & $\pm$0.22 & $\pm$2.33 \\
\hline\hline
\end{tabular}
\label{tab:app_perf_apbev}
\end{table*}

\noindent\textbf{Multiple Random Seeds.} Tab.~\ref{tab:app_perf_apbev} presents the mean and standard deviation of ASF (C+L+R) performance across 10 random seeds on the K-Radar benchmark v1.0, demonstrating the robustness and generality of our approach. The low standard deviations (mostly below 0.3) indicate stable performance across different random initializations. Notably, the sleet and heavy snow condition shows higher variance ($\pm$2.85 for $AP_{BEV}$ and $\pm$2.33 for $AP_{3D}$, respectively), suggesting this weather condition presents more challenging and variable detection scenarios. The consistently high performance in Fog conditions (98.97 $AP_{BEV}$ and 90.67 $AP_{3D}$) with minimal variance demonstrates ASF's particular effectiveness in low-visibility scenarios where 4D Radar excels.

\begin{table*}[h]
\caption{Performance comparison of monocular and stereo camera-based networks on the K-Radar \citep{kradar} benchmark v2.0. We report both $AP_{BEV}$ and $AP_{3D}$ at IoU = 0.3 for the `Sedan' and `Bus or Truck' classes. We note that camera performance is evaluated on sequences 1 to 20, which consist of normal weather conditions.}
\centering
\small
\begin{tabular}{c|c|cc|cc}
\hline\hline
Metric & Class & BEVDepth (Mono) & DSGN++ (Stereo) & ASF (BEVDepth) & ASF (Stereo) \\
\hline
\multirow{2}{*}{$AP_{BEV}$} & Sedan         & 19.9 & 24.3 & 82.5 & 82.4  \\
                            & Bus or Truck  & 13.4 & 24.8 & 70.2 & 67.2  \\
\hline
\multirow{2}{*}{$AP_{3D}$} & Sedan          & 17.8 & 20.8 & 79.4 & 79.3  \\
                            & Bus or Truck  & 11.5 & 22.5 & 60.4 & 59.5 \\
\hline\hline
\end{tabular}
\label{tab:app_stereo}
\end{table*}

\noindent\textbf{Stereo Camera.} As shown in Tab.~\ref{tab:app_stereo}, the stereo camera-based DSGN++ \citep{dsgnp} achieves improved performance on K-Radar (24.8\% and 22.5\% $AP_{3D}$ for Sedan and Bus or Truck, respectively) compared to the monocular camera-based BEVDepth \citep{bevdepth}. We integrate DSGN++ as ASF's camera-specific encoder. However, as shown in Tab.~\ref{tab:app_stereo}, ASF using DSGN++ shows comparable performance to ASF using BEVDepth. This may be attributed to two factors: (1) the detection performance of LiDAR and 4D Radar is significantly superior to that of the camera, and (2) K-Radar's ZED 2i camera has a 12cm baseline compared to KITTI's 54cm, limiting the improvement potential of stereo-based methods.

\noindent\textbf{Performance on Benchmark v2.0.} Tab.~\ref{tab:app_perf1}, \ref{tab:app_perf2}, \ref{tab:app_perf3}, and \ref{tab:app_perf4} present comprehensive evaluations of ASF performance across various sensor configurations and weather conditions on the K-Radar benchmark v2.0. Results demonstrate that multi-sensor fusion approaches consistently outperform single-sensor methods. The C+L+R configuration achieves the highest performance in most scenarios for both `Sedan' and `Bus or Truck' classes, with C*+L+R often ranking second, which indicates that a damaged camera does not significantly affect performance.

Performance analysis across weather conditions reveals that 4D Radar-inclusive combinations (C+L+R, L+R) exhibit superior performance in adverse weather (demonstrating 4D Radar's effectiveness under challenging environmental conditions), particularly in fog and snow. While camera-LiDAR fusion (C+L) occasionally outperforms in normal and overcast conditions, its performance deteriorates significantly in precipitation.

\begin{table*}[h]
\caption{{Performance comparison of ASF under various sensor combinations evaluated on the K-Radar \citep{kradar} benchmark v2.0.} We indicate the employed sensors (C: Camera, L: LiDAR, R: 4D Radar) and report $AP_{3D}$ at IoU = 0.3 under various weather conditions for the `Sedan' and `Bus or Truck' classes. The \textbf{bold} and \underline{underlined} values indicate the best and the second-best, respectively. All ten ASF models share the same neural network weights trained with R+L+C.}
\centering
\small
\begin{tabular}{c|c|c|ccccccccc}
\hline\hline
Class & Sensors & Total & Normal & Overcast & Fog & Rain & Sleet & Light snow & Heavy snow \\
\hline
\multirow{10}{*}{Sedan} 
  & R & 47.3 & 40.7 & 58.8 & 84.4 & 41.0 & 45.9 & 66.1 & 56.5 \\
  & L & 73.0 & 73.0 & 86.1 & 92.9 & 64.3 & 64.9 & 86.0 & 54.5 \\
  & C & 14.8 & 14.9 & 7.7 & - & - & - & - & - \\
  & C$^{*}$ & 3.7 & 3.7 & 3.2 & - & - & - & - & - \\
  & L+R & 77.3 & 77.7 & \underline{87.3} & \textbf{95.5} & 71.0 & \textbf{74.4} & 91.0 & \underline{65.4} \\
  & C+R & 52.7 & 49.1 & 62.4 & 84.6 & 44.6 & 46.0 & 66.5 & 57.2 \\
  & C+L & 76.4 & \underline{78.3} & 86.5 & 93.4 & 69.4 & 64.2 & 87.3 & 57.1 \\
  & C+L+R & \textbf{79.3} & \textbf{78.8} & 86.1 & 93.7 & \textbf{72.9} & \underline{74.2} & \textbf{91.2} & \textbf{65.8} \\
  & C$^{*}$+L+R & \underline{77.6} & 78.2 & \textbf{87.7} & \underline{93.8} & \underline{71.3} & \textbf{74.4} & \underline{91.1} & \underline{65.4} \\
  & C+L$^{*}$+R & 58.9 & 59.2 & 64.0 & 92.4 & 47.7 & 65.4 & 66.5 & 58.2 \\
\hline
\multirow{10}{*}{\makecell{Bus\\or\\Truck}} 
  & R & 34.2 & 22.9 & 41.0 & - & 0.2 & 21.1 & 83.5 & 51.2 \\
  & L & 54.9 & \textbf{53.7} & \underline{74.8} & - & 5.1 & 69.1 & 85.2 & 37.8 \\
  & C & 9.6 & 9.0 & 17.2 & - & - & - & - & - \\
  & C$^{*}$ & 3.7 & 3.7 & 0.0 & - & - & - & - & - \\
  & L+R & 59.9 & 52.5 & 71.6 & - & 6.8 & 68.2 & \textbf{88.1} & \underline{68.9} \\
  & C+R & 36.2 & 24.4 & 41.4 & - & 0.3 & 23.4 & 82.1 & 56.5 \\
  & C+L & 53.0 & 49.1 & 60.1 & - & 4.4 & \textbf{72.1} & 85.1 & 39.6 \\
  & C+L+R & \textbf{60.4} & \underline{52.7} & \textbf{77.4} & - & \underline{8.0} & 69.2 & \underline{87.9} & \underline{68.9} \\
  & C$^{*}$+L+R & \underline{60.1} & 52.1 & 72.0 & - & \textbf{8.4} & \underline{70.9} & \underline{87.9} & \textbf{69.1} \\
  & C+L$^{*}$+R & 40.0 & 31.5 & 38.2 & - & 0.4 & 28.9 & 81.0 & 54.8 \\
\hline\hline
\end{tabular}
\label{tab:app_perf1}
\end{table*}

\begin{table*}[h]
    \caption{{Performance comparison of ASF under various sensor combinations evaluated on the K-Radar \citep{kradar} benchmark v2.0.} We indicate the employed sensors (C: Camera, L: LiDAR, R: 4D Radar) and report $AP_{BEV}$ at IoU = 0.3 under various weather conditions for the `Sedan' and `Bus or Truck' classes. The \textbf{bold} and \underline{underlined} values indicate the best and the second-best, respectively. All ten ASF models share the same neural network weights trained with R+L+C.}
    \centering
    \small
    \begin{tabular}{c|c|c|ccccccccc}
        \hline\hline
        Class & Sensors & Total & Normal & Overcast & Fog & Rain & Sleet & Light snow & Heavy snow \\
        \hline
        \multirow{10}{*}{Sedan} 
                                & R     & 55.9 & 48.7 & 70.1 & 92.0 & 50.8 & 52.6 & 76.8 & 61.1 \\
                                & L     & 76.7 & 78.1 & 93.8 & 93.6 & 70.0 & 67.8 & 88.4 & 58.2  \\
                                & C     & 17.6 & 17.7 & 8.9 & - & - & - & - & - \\                                                           
                                & C$^{*}$ & 5.3 & 5.3 & 3.5 & - & - & - & - & - \\
                                & L+R   & 81.8 & 80.7 & 94.3 & \textbf{98.1} & 75.7 & \textbf{81.0} & \underline{93.6} & \textbf{69.6}  \\
                                & C+R   & 58.7 & 54.7 & 71.1 & 91.9 & 52.7 & 52.4 & 74.6 & 61.3  \\
                                & C+L   & 79.3 & 81.0 & \textbf{94.9} & 95.1 & 72.5 & 68.3 & 89.7 & 59.5  \\
                                & C+L+R & \textbf{82.5} & \textbf{81.8} & \textbf{94.9} & 98.0 & \textbf{76.2} & \textbf{81.0} & \textbf{93.8} & \underline{69.5}   \\
                                & C$^{*}$+L+R & \underline{82.1} & \underline{81.1} & \underline{94.7} & \textbf{98.1} & \underline{76.0} & 80.2 & \textbf{93.8} & 69.4 \\
                                & C+L$^{*}$+R & 64.5 & 64.6 & 76.1 & 92.5 & 57.2 & 56.9 & 75.9 & 60.6     \\
        \hline
        \multirow{10}{*}{\makecell{Bus\\or\\Truck}}
                               & R     & 43.5 & 28.4 & 42.9 & - & 0.3 & 23.7 & 92.5 & 77.2 \\
                               & L     & 63.1 & 57.0 & 75.5 & - & 6.2 & 72.3 & 88.7 & 61.3 \\
                               & C     & 11.4 & 11.2 & 18.1 & - & - & - & - & - \\
                               & C$^{*}$ & 3.9 & 4.0 & 0.0 & - & - & - & - & - \\
                               & L+R   & 69.5 & \textbf{58.3} & 75.2 & - & 7.3 & 71.3 & \underline{96.2} & \textbf{89.7} \\
                               & C+R   & 44.6 & 29.2 & 43.4 & - & 0.3 & 26.0 & 91.1 & 76.9   \\
                               & C+L   & 61.8 & 53.1 & 68.0 & - & 5.3 & \textbf{75.5} & 90.0 & 64.7  \\
                               & C+L+R & \textbf{70.2} & \textbf{58.3} & \textbf{81.5} & - & \underline{8.0} & 72.2 & 96.1 & \underline{89.6}   \\
                               & C$^{*}$+L+R & \underline{70.0} & \underline{58.2} & \underline{75.7} & - & \textbf{9.2} & \underline{73.8} & \textbf{96.4} & 88.4   \\
                               & C+L$^{*}$+R & 48.0 & 36.2 & 42.6 & - & 0.4 & 33.6 & 91.0 & 77.2   \\
        \hline\hline
    \end{tabular}
    \label{tab:app_perf2}
\end{table*}

\begin{table*}[h]
    \caption{{Performance comparison of ASF under various sensor combinations evaluated on the K-Radar \citep{kradar} benchmark v2.0.} We indicate the employed sensors (C: Camera, L: LiDAR, R: 4D Radar) and report $AP_{3D}$ at IoU = 0.5 under various weather conditions for the `Sedan' and `Bus or Truck' classes. The \textbf{bold} and \underline{underlined} values indicate the best and the second-best, respectively. All ten ASF models share the same neural network weights trained with R+L+C.}
    \centering
    \small
    \begin{tabular}{c|c|c|ccccccccc}
        \hline\hline
        Class & Sensors & Total & Normal & Overcast & Fog & Rain & Sleet & Light snow & Heavy snow \\
        \hline
        \multirow{10}{*}{Sedan} 
                               & R     & 20.4 & 13.2 & 26.6 & 59.5 & 18.1 & 20.9 & 28.7 & 30.7 \\
                               & L     & 41.1 & 38.5 & 43.1 & 67.8 & 37.1 & 34.1 & 52.4 & 35.4  \\
                               & C     & 5.9 & 5.9 & 2.8 & - & - & - & - & - \\                                                           
                               & C$^{*}$ & 1.0 & 1.0 & 0.8 & - & - & - & - & - \\
                               & L+R   & 52.1 & 49.0 & \textbf{56.9} & \underline{81.4} & 48.2 & \textbf{45.1} & 59.6 & \textbf{50.1}  \\
                               & C+R   & 27.5 & 22.6 & 29.8 & 66.2 & 23.1 & 27.2 & 30.2 & 39.3  \\
                               & C+L   & \underline{52.3} & \textbf{52.3} & 50.0 & 79.4 & 45.7 & 40.1 & \textbf{65.9} & 42.7  \\
                               & C+L+R & \textbf{52.9} & \underline{50.5} & 55.8 & \textbf{81.5} & \textbf{49.1} & 44.6 & 60.1 & \underline{50.0}   \\
                               & C$^{*}$+L+R & \underline{52.3} & 49.3 & \underline{56.8} & 79.6 & \underline{48.9} & \underline{45.0} & \underline{60.3} & \underline{50.0} \\
                               & C+L$^{*}$+R & 33.7 & 29.9 & 32.3 & 67.5 & 29.0 & 29.1 & 37.6 & 38.7 \\
        \hline
        \multirow{10}{*}{\makecell{Bus\\or\\Truck}}
                               & R     & 16.5 & 8.5 & 22.6 & - & 0.1 & 8.7 & 65.3 & 24.6 \\
                               & L     & 29.1 & 22.1 & 38.0 & - & 1.1 & 44.2 & 66.2 & 20.5 \\
                               & C     & 2.6 & 2.5 & 6.3 & - & - & - & - & - \\
                               & C$^{*}$ & 2.6 & 2.6 & 0.0 & - & - & - & - & - \\
                               & L+R   & \underline{31.7} & \underline{24.2} & \textbf{52.6} & - & \textbf{2.8} & 44.4 & \underline{70.9} & \underline{33.2} \\
                               & C+R   & 18.1 & 9.7 & 18.3 & - & 0.2 & 8.7 & 65.1 & 28.0   \\
                               & C+L   & 31.2 & 22.5 & 33.6 & - & 0.9 & \textbf{51.4} & 69.6 & 21.2  \\
                               & C+L+R & 31.6 & 23.7 & 47.0 & - & \underline{2.5} & \underline{45.0} & \textbf{72.3} & 33.0   \\
                               & C$^{*}$+L+R & \textbf{32.1} & \textbf{24.5} & \underline{48.2} & - & 2.3 & 44.8 & 70.6 & \textbf{33.3}   \\
                               & C+L$^{*}$+R & 19.2 & 13.5 & 15.6 & - & 0.0 & 9.8 & 62.5 & 27.9   \\
        \hline\hline
    \end{tabular}
    \label{tab:app_perf3}
\end{table*}

\begin{table*}[h]
    \caption{{Performance comparison of ASF under various sensor combinations evaluated on the K-Radar \citep{kradar} benchmark v2.0.} We indicate the employed sensors (C: Camera, L: LiDAR, R: 4D Radar) and report $AP_{BEV}$ at IoU = 0.5 under various weather conditions for the `Sedan' and `Bus or Truck' classes. The \textbf{bold} and \underline{underlined} values indicate the best and the second-best, respectively. All ten ASF models share the same neural network weights trained with R+L+C.}
    \centering
    \small
    \begin{tabular}{c|c|c|ccccccccc}
        \hline\hline
        Class & Sensors & Total & Normal & Overcast & Fog & Rain & Sleet & Light snow & Heavy snow \\
        \hline
        \multirow{10}{*}{Sedan} 
                                & R     & 40.8 & 30.0 & 54.7 & 85.6 & 36.9 & 39.6 & 65.2 & 48.4 \\
                                & L     & 66.8 & 65.7 & 79.0 & 90.7 & 60.1 & 54.6 & 83.4 & 52.7  \\
                                & C     & 8.8 & 8.9 & 4.3 & - & - & - & - & - \\                                                           
                                & C$^{*}$ & 2.1 & 2.1 & 1.1 & - & - & - & - & - \\
                                & L+R   & 74.0 & 72.6 & 84.4 & \underline{96.1} & \textbf{67.3} & \underline{67.4} & \textbf{90.8} & \underline{63.4}  \\
                                & C+R   & 45.8 & 39.7 & 57.2 & 87.4 & 40.7 & 41.6 & 66.7 & 53.9  \\
                                & C+L   & 72.3 & 72.8 & \textbf{87.2} & 93.3 & 62.5 & 55.4 & 87.2 & 55.9  \\
                                & C+L+R & \textbf{74.5} & \textbf{74.8} & \underline{85.1} & \textbf{96.2} & \underline{67.0} & \textbf{67.5} & \textbf{90.8} & \textbf{63.5}   \\
                                & C$^{*}$+L+R & \underline{74.1} & \underline{72.9} & 85.0 & \underline{96.1} & 66.9 & 66.7 & 90.7 & 63.2 \\
                                & C+L$^{*}$+R & 53.1 & 54.5 & 57.0 & 61.5 & 50.2 & 50.3 & 68.5 & 55.2     \\
        \hline
        \multirow{10}{*}{\makecell{Bus\\or\\Truck}}
                               & R     & 23.8 & 14.2 & 30.0 & - & 0.2 & 12.7 & 87.3 & 64.3 \\
                               & L     & 45.8 & 35.7 & \textbf{61.1} & - & 2.3 & 58.2 & 81.9 & 53.3 \\
                               & C     & 11.4 & 11.2 & 18.1 & - & - & - & - & - \\
                               & C$^{*}$ & 2.7 & 2.8 & 0.0 & - & - & - & - & - \\
                               & L+R   & 53.1 & \textbf{37.5} & \underline{59.8} & - & 6.1 & 56.6 & 88.2 & \underline{80.1} \\
                               & C+R   & 31.9 & 14.4 & 25.9 & - & 0.2 & 13.4 & 90.1 & 64.7   \\
                               & C+L   & 47.3 & 32.4 & 40.7 & - & 3.3 & \textbf{63.1} & 85.3 & 55.6  \\
                               & C+L+R & \textbf{53.9} & \underline{36.5} & 54.7 & - & \underline{7.9} & 59.3 & \underline{90.4} & \textbf{80.4}   \\
                               & C$^{*}$+L+R & \underline{53.6} & 36.1 & 57.5 & - & \textbf{8.0} & \underline{59.7} & \textbf{90.8} & \underline{80.1}   \\
                               & C+L$^{*}$+R & 33.2 & 20.9 & 24.7 & - & 0.2 & 18.3 & 88.0 & 61.4   \\
        \hline\hline
    \end{tabular}
    \label{tab:app_perf4}
\end{table*}

\section*{Acknowledgment}

This work was supported by the National Research Foundation of Korea (NRF) grant funded by the Korea government (MSIT) (No. 2021R1A2C3008370). 

\bibliography{neurips_2025}

\end{document}